\documentclass[%
 aip,
 amsmath,amssymb,
 reprint,%
]{revtex4-1}

\usepackage{graphicx}
\usepackage{dcolumn}
\usepackage{bm}
\usepackage[utf8]{inputenc}
\usepackage[T1]{fontenc}
\usepackage{mathptmx}
\usepackage{etoolbox}
\usepackage{multirow}
\usepackage{diagbox}

\newcommand{\etal}{\textit{et al.}}

\usepackage{tikz}

\makeatletter
\def\@email#1#2{%
 \endgroup
 \patchcmd{\titleblock@produce}
  {\frontmatter@RRAPformat}
  {\frontmatter@RRAPformat{\produce@RRAP{*#1\href{mailto:#2}{#2}}}\frontmatter@RRAPformat}
  {}{}
}%
\makeatother
\begin{document}

\preprint{AIP/123-QED}

\title[When In-memory Computing Meets Spiking Neural Networks]{When In-memory Computing Meets Spiking Neural Networks $-$\\A Perspective on Device-Circuit-System-and-Algorithm Co-design}


\author{Abhishek Moitra}

\altaffiliation[]
 {These authors have contributed equally.}
 \affiliation{ 
$^1$Department of Electrical Engineering, Yale University, New Haven, CT, USA, 06511
}%
\author{Abhiroop Bhattacharjee}

\altaffiliation[]
 {These authors have contributed equally.}
 \affiliation{ 
$^1$Department of Electrical Engineering, Yale University, New Haven, CT, USA, 06511
}%
\author{Yuhang Li}
\affiliation{ 
$^1$Department of Electrical Engineering, Yale University, New Haven, CT, USA, 06511
}%
\author{Youngeun Kim}
\affiliation{ 
$^1$Department of Electrical Engineering, Yale University, New Haven, CT, USA, 06511
}%
\author{Priyadarshini Panda}
 
 \affiliation{ 
$^1$Department of Electrical Engineering, Yale University, New Haven, CT, USA, 06511
}%
 \email{\{abhishek.moitra, abhiroop.bhattacharjee, yuhang.li, youngeun.kim, priya.panda\}@yale.edu}

\date{\today}

\begin{abstract}
This review explores the intersection of bio-plausible artificial intelligence in the form of Spiking Neural Networks (SNNs) with the analog In-Memory Computing (IMC) domain, highlighting their collective potential for low-power edge computing environments. Through detailed investigation at the device, circuit, and system levels, we highlight the pivotal synergies between SNNs and IMC architectures. Additionally, we emphasize the critical need for comprehensive system-level analyses, considering the inter-dependencies between algorithms, devices, circuit \& system parameters, crucial for optimal performance. An in-depth analysis leads to identification of key system-level bottlenecks arising from device limitations which can be addressed using SNN-specific algorithm-hardware co-design techniques. This review underscores the imperative for holistic device to system design space co-exploration, highlighting the critical aspects of hardware and algorithm research endeavors for low-power neuromorphic solutions.
\end{abstract}

\maketitle


\section{Introduction}
Artificial Intelligence (AI) has been at the forefront of technological innovation over the past decade. From the development of deep convolutional neural networks \cite{krizhevsky2012imagenet, he2016deep} that have revolutionized computer vision to the emergence of transformers\cite{dosovitskiy2020image} and large language models\cite{devlin2018bert} that have transformed natural language processing, each generation of AI algorithm represents a significant leap in our ability to harness the power of data.
The growth of AI is mainly attributed to the scale-up of high power, server-class computing machines such as graphics processing units (GPUs) \cite{reuther2022ai}. But, GPUs draw a substantial amount of power, leading to increased operational costs and a larger carbon footprint \cite{kandiah2021accelwattch, reuther2022ai}.
 As AI aims to become more ubiquitous and user-centric, there is a growing need for low-power AI algorithms and hardware accelerators. This shift is essential to move away from the current trend of expecting increased intelligence merely by scaling up compute/memory resources, which is neither sustainable nor practical for widespread deployment. However, this vision stands in contrast to the current algorithm and hardware progress trajectory, where the computational needs of AI algorithms are doubling every two months, far surpassing Moore's law of silicon scaling by a considerable margin \cite{mehonic2022brain}. 


To this end, neuromorphic computing algorithms such as Spiking Neural Networks (SNNs) leveraging brain-like computations have emerged as a suitable candidate towards low-power AI implementation \cite{roy2019towards, kim2021revisiting, kim2020spiking}. 
An SNN contains numerous Leaky-Integrate-and-Fire (LIF) neurons that store the spatio-temporal information in the form of membrane potential values over multiple timesteps. The information from one neuron is relayed to another neuron in the form of binary spikes. Overall, due to the sparse event driven binary processing capabilities, SNNs show promise for low power edge computing applications, such as, in-sensor processing \cite{shaaban2024rt, maclean2024tdc, zhou2023computational, barchi2021spiking}, embedded intelligence \cite{li2023efficient, bian2023evaluating, tanzarella2023neuromorphic, gong2023spiking, fang2019neuro, qi2023neuromorphic}, among others. In fact, SNNs are increasingly being embraced by different industries for commercial products in image classification \cite{akopyan2015truenorth}, optimization \cite{davies2018loihi}, agriculture \cite{andante}, and autonomous driving \cite{tsst}, signaling widespread adoption and innovation potential. In recent years, SNNs have achieved comparable accuracy with standard Artificial Neural Networks (ANNs) in large-scale tasks such as image classification on the Imagenet-1K dataset \cite{deng2009imagenet, li2024seenn}. This has necessitated SNNs to scale up in terms of parameter count requiring significant compute and memory resources. Thus, their implementation on the emerging low-power hardware acceleration paradigm, In-Memory Computing (IMC), shows great promise as IMC facilitates highly parallel computation with high memory bandwidth.

Traditional von-Neumann style accelerators such as Graphics Processing Units (GPUs) and Tensor Processing Units (TPUs) suffer from "memory wall bottleneck" owing to the heavy data movement (specifically, weights) between memory and compute units corresponding to the dot-product operations in neural networks \cite{jouppi2017datacenter, xu2018deep}. IMC with non-volatile memories facilitates analog dot-product operations while keeping the weights stationary on crossbars, thereby reducing the weight movement bottleneck significantly \cite{verma2019memory, sebastian2020memory}. Further, the costly digital multiplier-accumulator circuits of von-Neumann accelerators are reduced to analog crossbar operation on IMC leading to energy and area efficiency. Importantly, SNNs exhibit tight synergies with IMC hardware. Due to the high spike sparsity (90\% across different layers \cite{yin2022sata}) and binary spike computations, SNNs implemented on IMC require low peripheral and data communication overhead that yield low-power and high-throughput benefits \cite{shanbhag2022comprehending, mehonic2022brain}.
\begin{figure*}
    \centering
    \includegraphics[width=0.8\linewidth]{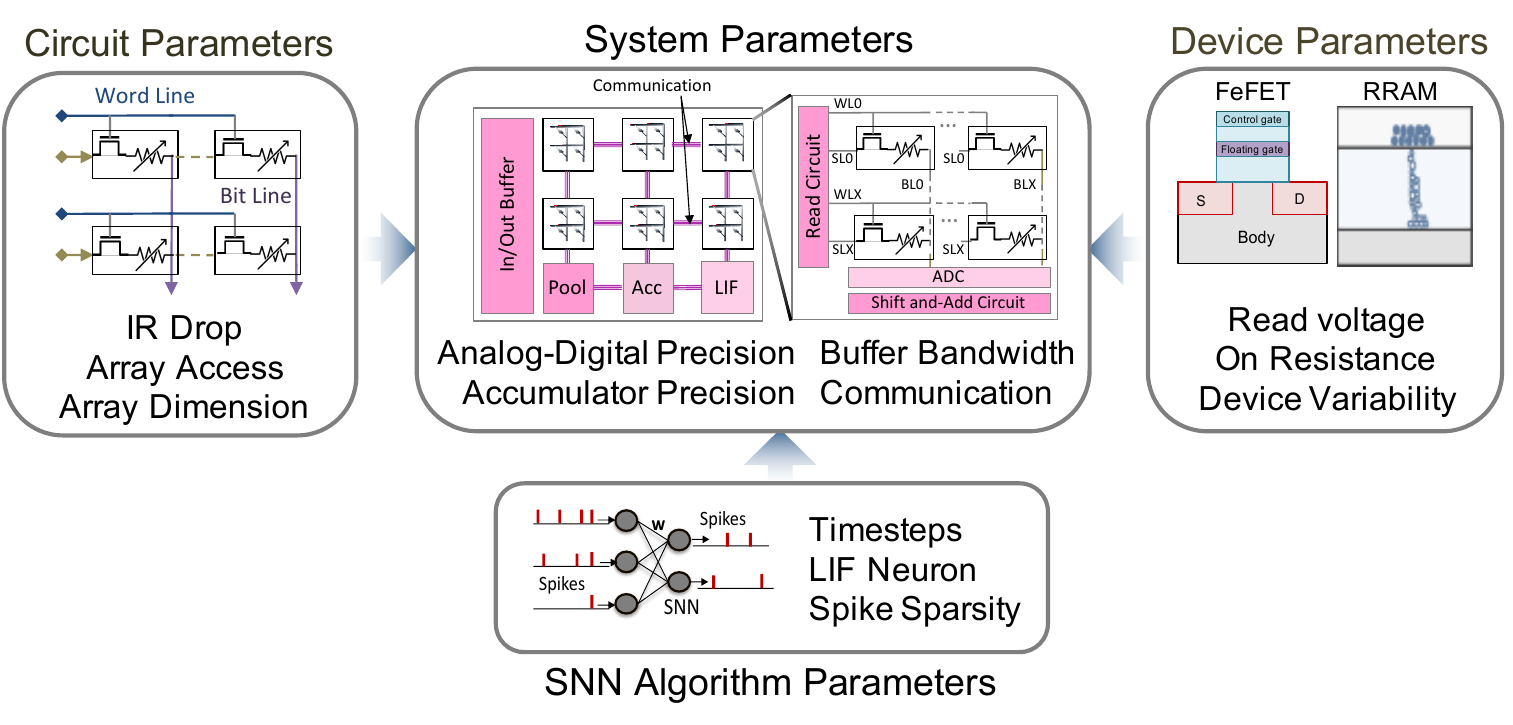}
    \caption{Landscape of the Spiking Neural Network (SNN) algorithm, In-memory computing (IMC) device, circuit and system parameters. In order to reach fully optimal SNN-IMC implementations, there is a critical need to consider the existing co-dependencies between IMC device, circuit, system and SNN algorithm parameters. LIF denotes leaky-integrate and fire neuron, a fundamental non-linear activation unit in SNNs. Relevant parameters for each domain that underlie hardware metrics such as, performance, latency, energy efficiency, area and power are mentioned.}  \vspace{-3mm}
    \label{fig:intro_fig}
   
\end{figure*}

Fig. \ref{fig:intro_fig} illustrates the prevailing landscape of research endeavors spanning device, circuit, and algorithm within the realm of SNN research. On the algorithmic front, scholars have directed their effort towards harnessing spatio-temporal computation, exploiting biological LIF neuron functionality, and optimizing spike sparsity to enhance the efficiency of SNN algorithms \cite{kim2021revisiting, kim2022exploring, kim2022neural}. Meanwhile, within the domain of devices and circuits, the community has prioritized objectives such as achieving multi-level conductances, mitigating device vulnerabilities, and optimizing crossbar connections, among other pursuits \cite{rao2023thousands, bhattacharjee2021neat, bhattacharjee2023switchx, moitra2021detectx}. However, to date, these research endeavors have largely been independent of one another. There exists a critical need to integrate various device, circuit, and algorithmic parameters with system-level considerations to realize truly optimal solutions. A comprehensive system-level inquiry holds the potential to bridge the gap between device-circuit and algorithmic-level innovations. In pursuit of this objective, the present review delineates the key bottlenecks and opportunities associated with implementing SNNs on emerging IMC architectures. With this, the review furnishes strategic guidance to diverse research communities on innovations for optimal implementation of SNNs on IMC architectures. 

In this review, we first discuss the algorithmic efficiencies and current state-of-the-art training algorithms for SNNs. Additionally, we highlight the widespread application space of SNNs. Thereafter, we highlight the existent synergies between SNNs and IMC architectures that make them a suitable candidate for low power edge computing. Following this, we motivate the need for end-to-end device, circuit and system-level analyses to understand the challenges of implementing SNNs on IMC platforms. We will also highlight the recent SNN-aware co-design strategies that can overcome the pressing bottlenecks. Finally, we highlight the key questions that lie ahead in the SNN-IMC research. We discuss the device and system-level parameters that are crucial towards SNN inference with future emphasis on sustainable online-learning.




\section{SNN Algorithm and Application Space}
\subsection{Inherent Efficiencies in SNNs}

\begin{figure}[]
    \centering
    \includegraphics[width=0.85\linewidth]{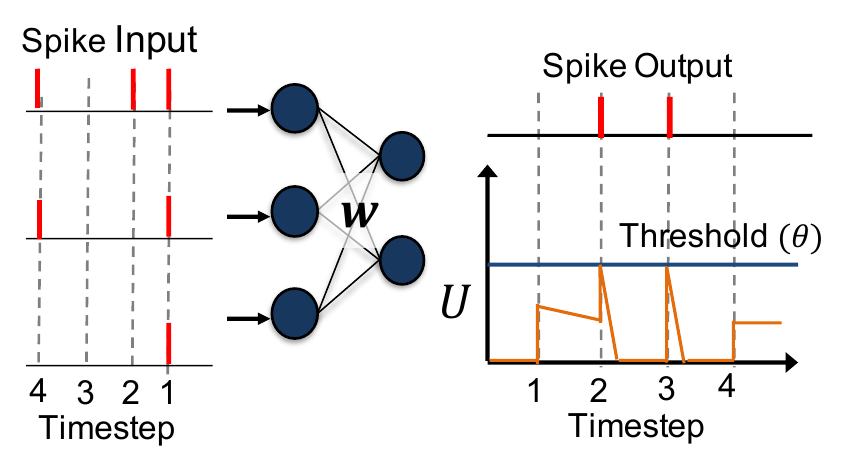}
    \caption{Figure showing the functioning of an SNN. Input spikes are sent to the SNN across multiple timesteps. These binary spikes get multiplied with SNN weights ($w$) to generate Multiply-and-Accumulate (MAC) values which charge the membrane potential value $U$ over multiple timesteps. At any timestep, if the membrane potential exceeds a pre-defined threshold ($\theta$), a spike output is generated.}
    \label{fig:snn_basics}
\end{figure}
SNNs inherently possess several key efficiencies that are critical towards low power edge implementation.

\textbf{(1) Binary Spike Processing:} Taking cues from the brain, SNNs perform binary spike-driven data processing over multiple timesteps. As a result, Multiply-and-Accumulate (MAC) operations are merely reduced to efficient accumulation operations \cite{yin2022sata, yin2023mint}.
    
\textbf{(2) Spatio-temporal Complexity:} At each timestep, the input spikes and SNN weights undergo spatial convolution yielding dot-product outputs. SNNs use a special non-linearity function called Leaky-integrate and Fire (LIF)\cite{kim2021revisiting, kim2022exploring}. {The dynamics of an LIF neuron is shown in Equation \ref{eq:LIF}. For neuron $i$, the membrane potential $U$ is charged by the weighted summation of spikes from the previous layer's neuron $j$ at every timestep $t$.} Also, the leak factor $\lambda \in (0,1)$ facilitates temporal leakage of the membrane potential. 
\begin{equation}
    U_i^t = \lambda  U_i^{t-1} + \sum_j w_{ij}o^t_j.
    \label{eq:LIF}
\end{equation}
If at any timestep, the value $U$ exceeds a particular threshold $\theta$, the neuron generates a spike output and vice-versa as shown in Equation \ref{eq:firing}.
\begin{equation}
    o^{t}_i =
    \begin{cases}
     1,          & \text{if $U_i^{t} >\theta$},  \\
        0
        & \text{otherwise.} 
    \end{cases}
    \label{eq:firing}
\end{equation}

This has been described in Fig. \ref{fig:snn_basics}. It is worth noting that the number of timesteps for processing the neuronal dynamics will eventually determine the overall performance of an SNN \cite{moitra2023spikesim, li2023input}. Generally, SNNs with high timesteps yield better accuracy than that of SNNs with less timesteps. But, larger timesteps also translate to higher latency and energy consumption on hardware. As we will see later (in Fig. \ref{fig:ts_overhead}), timesteps become a critical control knob to determine the overall energy-vs.-accuracy tradeoff while designing SNNs.

\textbf{(3) Data Sparsity:} The LIF neuron activation yields high spike sparsity. This means that SNNs can represent data with very few spikes. At any given timestep, around 90\% of the neurons in an SNN are not spiking. Compared to ANNs with the ReLU activation, SNNs exhibit at least 30-40\% higher neuronal sparsity \cite{kim2022exploring, kim2021revisiting}. 

\textbf{(4) Event-driven Computation:} Due to the high spike sparsity, leveraging event-driven computation and communication can significantly improve the energy-efficiency of hardware accelerators \cite{narayanan2020spinalflow, yin2022sata, ankit2017resparc}.

\begin{figure*}
    \centering
    \includegraphics[width=0.9\linewidth]{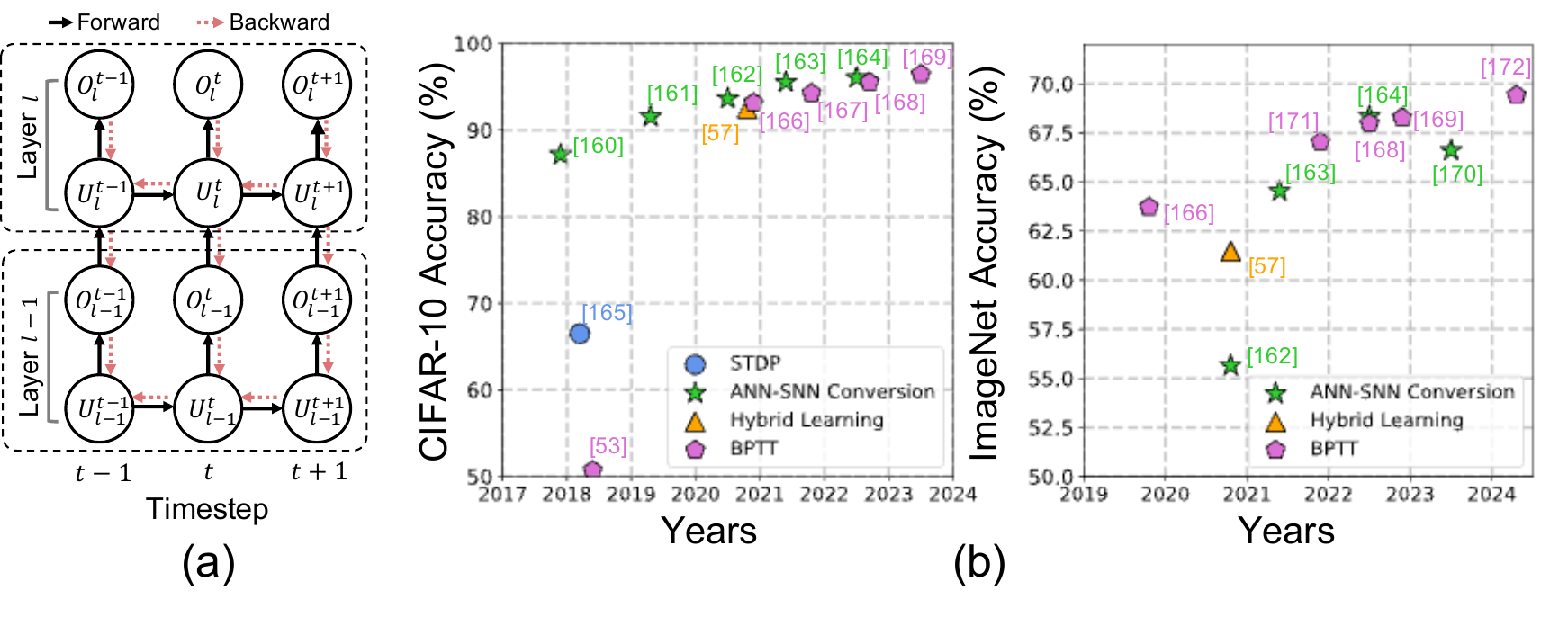}
    \caption{(a) Training computation graph for the BPTT algorithm. The gradients pass through different layers and timesteps. (b) SNN accuracy on image classification tasks over the years. We show accuracy on two widely used benchmarks: CIFAR10 \cite{cifar10} and ImageNet \cite{deng2009imagenet}, that comprise of 50000, 1.2 Million images with 10 classes and 1000 classes, respectively. {SNNs trained on BPTT can achieve high accuracy while scaling to large-scale datasets at low timesteps}. }
    \label{fig:algorithm}
\end{figure*}

\subsection{State-of-the-art SNN Training Algorithms}
\label{sec:Algorithm}

    


In this section, we cover SNN training algorithms, including conventional learning algorithms like unsupervised Hebbian learning or Spike Timing-Dependent Plasticity (STDP),  ANN-SNN Conversion, and modern learning algorithms like Backpropagation Through Time (BPTT). 
We will discuss the scalability and practicality of each algorithm.

\subsubsection{Conventional Learning Algorithms}
\textbf{STDP Learning: } The Spike Timing-Dependent Plasticity~\cite{caporale2008spike} can be viewed as a spike-based formulation of a Hebbian learning rule~\cite{munakata2004hebbian}, where the synaptic weight is updated based on the tight temporal correlation of the firing time of the input and the output spikes.  
With STDP, if the presynaptic neuron fires briefly before the postsynaptic neuron (\textit{i.e.,} if the output spike timing is shortly after the input spike timing), the weight connecting them is strengthened. Otherwise, if the timing is reversed, the weight connection is weakened. 
This strengthening and weakening is called Long-Term Potentiation (LTP) and Long-Term Depression (LTD), respectively. 
STDP is a fully unsupervised learning algorithm that does not need a loss objective to update the weight neurons. Therefore, the SNNs trained with STDP are usually limited to thousands of neurons which limits the scalability of of this method~\cite{masquelier2007unsupervised, lee2018deep}. 
Recent advances~\cite{lee2018training} propose to incorporate STDP with supervised learning to increase their scalability to complex tasks.

\textbf{ANN-SNN Conversion: }
Given that ANN training is easier than SNN training, a straightforward way to obtain the SNN is to first train an ANN with the same architecture and convert the neurons into spiking neurons. The conversion process typically involves finding the optimal threshold values ($\theta$ in Eq. \ref{eq:firing}) for the membrane potential of the spiking neurons and scaling of the weights such that the spike rate in SNNs match the floating-point outputs of ANNs  \cite{cao2015spiking, diehl2015fast, han2020deep}. {In a different ANN-SNN conversion method, Han \etal \cite{han2020rmp} achieved improved convergence and higher accuracy for converted SNNs by performing ANN-SNN conversion without resetting the LIF neuron.}
The ANN-SNN conversion shares several advantages compared to direct training of SNNs. For example, ANN training is easy to implement since the computing hardware (e.g., GPUs) and the deep-learning library (e.g., PyTorch) are well-established. In addition, the conversion process is also simple as it only involves changing the neuron type of the model. 
However, this method also incurs several disadvantages: (1) The converted SNN requires significantly larger timesteps to realize the original ANN performance \cite{kim2021revisiting}. Large timesteps will translate to high latency or energy consumption on hardware implementation; (2) The converted SNN shares the same architecture with the ANNs, which is not tailored for the spike-based mechanism and does not fully utilize the temporal dynamics of SNNs~\cite{kim2022neural}.

\subsubsection{Back Propagation Through Time}


Previous methods like conversion or unsupervised learning suffer from either large timesteps or low scalability. The BPTT algorithm can address these two challenges. BPTT usually trains an SNN from scratch where the gradients are computed in a timestep-unrolled computation graph (see Fig. \ref{fig:algorithm}a). 
Formally, the gradient of the weight $w_{ij}$, denoted by $\Delta w_{ij}$, is accumulated over $T$ timesteps as:
\begin{equation}
    \Delta w_{ij} = \sum_{t = 1}^{T} \frac{\partial \mathcal{L}}{\partial o_{i}^{t}} \frac{\partial o_{i}^{t}}{\partial U_{i}^{t}} \frac{\partial U_{i}^{t}}{\partial w_{ij}},
\end{equation}
where, $\mathcal{L}$ is the loss function being optimized. In the case of image classification, categorical cross-entropy loss is widely used.
The challenge of applying gradient descent in SNNs is that the spike function returns zero gradient almost everywhere because of the thresholding function. Surrogate gradient descent \cite{neftci2019surrogate,wu2018spatio,wu2019direct,lee2016training, shaban2021adaptive} overcomes this problem by approximating the spike function into piece-wise linear, fast sigmoid or exponential function \cite{snn_bp}. For instance, the surrogate gradient descent method using a piece-wise linear approximation is defined as:
\begin{equation}\label{surrogate_gd}
    \frac{\partial o_i^t}{\partial U_i^t} = \xi \max \{0, 1-  \ | \frac{U_i^t - \theta}{\theta} \ | \},
\end{equation}
where, $\xi$ is a decay factor for back-propagated gradients and $\theta$ is the threshold value.
The hyperparameter $\xi$ should be set based on the total number of timesteps $T$. 
BPTT-based SNN reaches state-of-the-art accuracy on various tasks such as image recognition and event data processing at fewer timesteps.
Hybrid training \citep{rathi2020enabling} combines BPTT and ANN-SNN conversion in order to achieve higher accuracy. 
In Fig. \ref{fig:algorithm}b, we illustrate the accuracy of various SNN training methods over the last decade. {Evidently, BPTT-based training exhibits high accuracy while scaling to large-scale datasets such as Imagenet at low timestep overhead.} 

\subsection{Application Space for SNNs}

This section reviews the recent academic studies and commercial application space of SNNs (summarized in Table \ref{tab:snn_applications}).

\begin{table}
\Huge
    \centering
    \resizebox{\linewidth}{!}{%
    \begin{tabular}{|c|cc|} \hline
        \multirow{12}{*}{\rotatebox{90}{\textbf{Academic Studies}}} 
        & \multirow{1}{*}{\begin{tabular}{c} Efficient Sensing \end{tabular}} 
        & \begin{tabular}{c}
            Radar Data Inference \cite{shaaban2024rt}\\ 
            Depth Estimation \cite{maclean2024tdc}\\ 
            In-sensor Processing \cite{zhou2023computational, barchi2021spiking}
          \end{tabular} \\ \cline{2-3}
        
        & \multirow{1}{*}{RGB Cameras} 
        & \begin{tabular}{c}
            Object Detection \cite{kim2020spiking, su2023deep}\\ 
            Automotive Artifact Detection \cite{lopez2021spiking}
          \end{tabular} \\ \cline{2-3}

        & \multirow{1}{*}{Event Cameras} 
        & \begin{tabular}{c}
            Satellite Detection \cite{salvatore2023dynamic}\\ 
            Optical Flow Estimation \cite{yang2023sa, zheng2022spike}\\
            Gesture Recognition \cite{amir2017low, maro2020event, vasudevan2020introduction}
          \end{tabular} \\ \cline{2-3}

        & {\begin{tabular}{c}
            High-speed \\ Spike Cameras
          \end{tabular}} 
        & \begin{tabular}{c}
            Optical Flow Estimation \cite{zhai2023spike, chen2023self, xia2023unsupervised}\\ 
            Object Tracking \cite{zhao2023spireco, zheng2022spike}
          \end{tabular} \\ \cline{2-3}

        & {Wearable Healthcare} 
        & \begin{tabular}{c}
            Human Activity Recognition \cite{li2023efficient, bian2023evaluating, tanzarella2023neuromorphic}\\ 
            Anomaly Detection \cite{dalgaty2024mosaic, tian2023neurocare}\\
            Flexible Electronics \cite{wang2023neuromorphic}\\ 
            Brain-Computer Interface \cite{gong2023spiking, fang2019neuro, qi2023neuromorphic}
          \end{tabular} \\ \hline

        \multirow{7}{*}{\rotatebox{90}{\begin{tabular}{c}
        \textbf{Commercial} \\ 
        \textbf{Applications}
        \end{tabular}}} 
        & Image Classification 
        & IBM TrueNorth \cite{akopyan2015truenorth} \\ \cline{2-3}

        & Optimization 
        & Intel Loihi \cite{davies2018loihi} \\ \cline{2-3}

        & SLAM \& Odometry 
        & GrayScale-AI \cite{grayscaleai} \\ \cline{2-3}

        & Agriculture \& Healthcare 
        & Andante \cite{andante} \\ \cline{2-3}

        & Autonomous Driving 
        & TSST \cite{tsst} \\ \cline{2-3}

        & General Intelligence 
        & ORBAI \cite{orbai} \\ \hline
    \end{tabular}}
    \caption{Table showing the academic and industry application space of SNNs.}
    \label{tab:snn_applications}
\end{table}

\textbf{In-sensor Processing and Low Power Healthcare:} Due to their inherently sparse and binary spike representation, SNNs effectively reduce the bandwidth requirements for inter-chip interfacing. Works by Shaaban \etal \cite{shaaban2024rt} and MacLean \etal \cite{maclean2024tdc} employ efficient time-domain processing to replace computationally intensive preprocessing steps, while Zhou \etal \cite{zhou2023computational} and Barchi \etal \cite{barchi2021spiking} directly interface sensors with SNNs for in-sensor processing. 

The low power nature of SNNs have also been leveraged in wearable healthcare devices. NeuroCARE \cite{tian2023neurocare} and Mosaic \cite{dalgaty2024mosaic} offer tailored neuromorphic healthcare frameworks. Bian \etal \cite{bian2023evaluating} and Li \etal \cite{li2023efficient} utilize SNNs for human activity recognition in wearables, and Tanzarella \etal \cite{tanzarella2023neuromorphic} detect spinal motor neuron activity. Additionally, SNNs in Brain-Computer Interface (BCI) applications, explored by Gong \etal \cite{gong2023spiking} and Feng \etal \cite{feng2023towards}, leverage their energy-efficiency for Electroencephalography (EEG) analysis.

\textbf{Emergence of Event-driven and Spike Cameras:} Recent research in emerging vision cameras has broadened the applicability of SNNs, including object detection tasks. Initiatives like Spiking YOLO \cite{kim2020spiking} pioneered ANN-SNN conversion, enhancing object detection efficiency across various datasets. Subsequent works, like Su \etal \cite{su2023deep}, achieved state-of-the-art object detection via full-scale SNN training. In automotive applications, Lopez \etal \cite{lopez2021spiking} effectively utilized SNNs' sparse data representations for artifact detection. Salvatore \etal \cite{salvatore2023dynamic} demonstrated SNN robustness against noise and their effectiveness in satellite detection using event cameras. {Amir \etal \cite{amir2017low} Maro \etal \cite{maro2020event} and Vasudevan \etal \cite{vasudevan2020introduction} have performed gesture recognition using event cameras. While Amir \etal \cite{amir2017low} implement their algorithm on the event-driven TrueNorth \cite{akopyan2015truenorth} neuromorphic processor, Maro \etal \cite{maro2020event} implement their algorithm on an android smartphone. The DvsGesture, NavGesture and SL-Animals-DVS datasets proposed by Amir \etal \cite{amir2017low}, Maro \etal \cite{maro2020event} and Vasudevan \etal \cite{vasudevan2020introduction}, respectively can serve as temporal  datasets for benchmarking SNNs.} Moreover, studies such as Yang \etal \cite{yang2023sa} and Zheng \etal \cite{zheng2022spike} leveraged event cameras' spike-driven nature for optical flow estimation.

Recent studies explore advanced spike cameras due to limitations of conventional RGB and event cameras. Works by Zhai \etal \cite{zhai2023spike} and Chen \etal \cite{chen2023self} utilize SNNs for optical flow estimation and image reconstruction, showcasing spike cameras as the new standard in vision sensing technology. Additionally, Xia \etal \cite{xia2023unsupervised}, Zhai \etal \cite{zhai2023spike}, and Chen \etal \cite{chen2023self} demonstrate unsupervised SNN training for spike camera applications, while Zhao \etal \cite{zhao2023spireco} emphasize spike cameras' potential for high-speed object tracking, indicating ongoing advancements in spike camera technology.

\textbf{Industry Adoption of SNNs:} Industry leaders are increasingly embracing SNN-based solutions for product design across diverse sectors. IBM utilizes SNNs for image classification and detection through their TrueNorth \cite{akopyan2015truenorth} platform. Intel leverages its Loihi chip \cite{davies2018loihi} for optimization tasks, harnessing the power of SNNs. GrayScale-AI \cite{grayscaleai} employs SNNs for Simultaneous Localization and Mapping (SLAM) and Visual Odometry applications, while Andante \cite{andante} focuses on edge computing solutions for agriculture. In the healthcare sector, Andante \cite{andante} is employing SNNs for glucose monitoring and ultrasound image processing. TSST \cite{tsst} integrates SNNs with Ferroelectric Field-Effect Transistors (FeFETs) for autonomous driving applications, enhancing safety and efficiency. ORBAI \cite{orbai} capitalizes on SNNs to emulate human brain characteristics, such as associativity and problem-solving, in the development of Artificial General Intelligence (AGI). Notably, companies like Apple and Mercedes are also exploring the potential of SNNs in various product design initiatives, signaling a widespread adoption of this innovative technology across industries.

\section{IMC Accelerators for SNNs}
\subsection{von-Neumann and IMC Accelerators}
In this section, we discuss the differences between von-Neumann and IMC architectures for inference applications. Note, the discussion in Section III and Section IV will focus on inference, and Section V will highlight some opportunities and challenges for training with IMC-SNNs.
\begin{figure*}[t]
    \centering
    \includegraphics[width=0.9\linewidth]{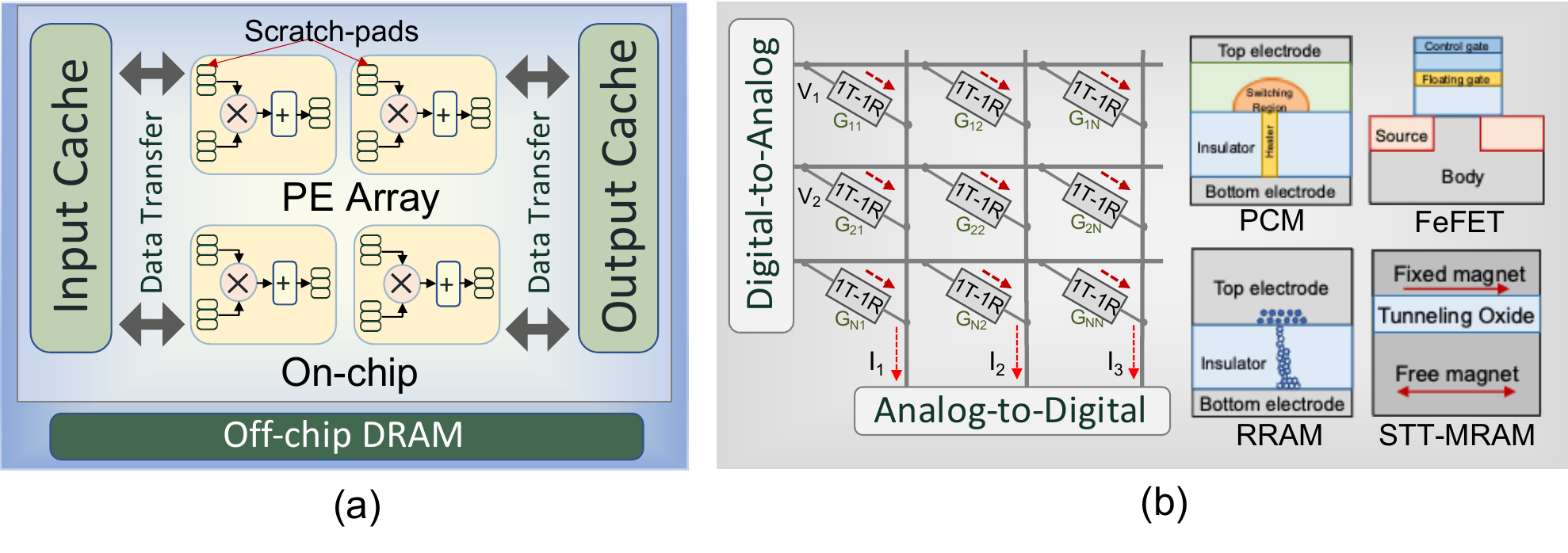}
    \caption{Figure showing (a) von-Neumann accelerators containing on-chip cache, scratch pad memories, multipliers and accumulators for performing MAC operations. (b) IMC architectures containing 2D arrays of 1 transistor-1 memsistor (1T-1R) devices. They perform fast analog dot-products minimizing data transfer to mitigate the ``memory wall bottleneck" typical in von-Neumann architectures. Over the years, different non-volatile memory (NVM) devices like Phase change memory (PCM), ferro-electric field effect transistor (FeFET), resistive random access memory (RRAM) and spin torque transfer-based magnetic RAM (STT-MRAM) have been used as memristors.}
\label{fig:vn_imc}
\vspace{-2mm}
\end{figure*}

\begin{figure*}[t]
    \centering
    \includegraphics[width=1\linewidth]{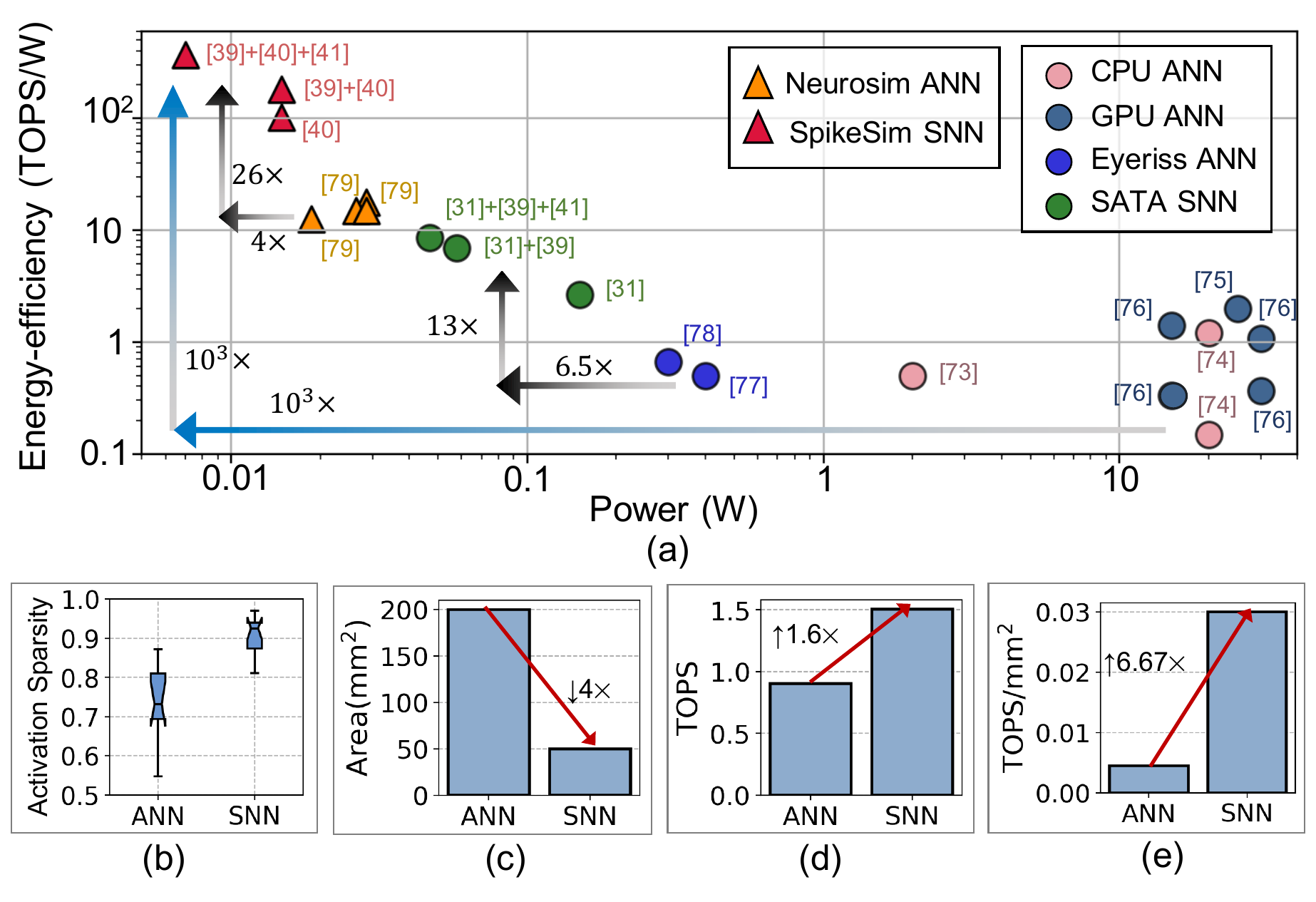}

    \caption{(a) Plot of energy-efficiency (measured in TOPS/W) vs Power for different low power edge AI accelerators. ANN workloads are deployed on CPU (Intel Movidius \cite{intel_movidius}, Kalray \cite{kalray}), GPU (Nvidia Jetson Orin Nano \cite{nvidia_orin_nano}, Nvidia Xavier \cite{nvidia_xavier}), systolic accelerators (Eyeriss-V1 \cite{chen2016eyeriss}, Eyeriss-V2\cite{chen2019eyeriss}), and IMC (Neurosim \cite{chen2018neurosim}) accelerators. SNNs are deployed on SATA \cite{yin2022sata} systolic array and SpikeSim \cite{moitra2023spikesim} IMC accelerator platforms. SATA \cite{yin2022sata} and SpikeSim\cite{moitra2023spikesim} are SNN-specific accelerators that closely resemble the Eyeriss \cite{chen2016eyeriss} and Neurosim\cite{chen2018neurosim} platforms, respectively and thus, facilitate a fair comparison. ``$+$" denotes the conjunction of two approaches. Arrows are used to show the reduction and improvements in power and energy-efficiency, respectively. Higher energy-efficiency at lower power signifies a good AI accelerator platform. (b) Plot comparing the activation sparsity averaged across different layers of the ANN and SNN. Plots comparing (c) IMC chip area (d) TOPS and (e) TOPS/mm$^2$ of ANN and SNN implemented on the Neurosim \cite{chen2018neurosim} and SpikeSim \cite{moitra2023spikesim} platforms, respectively. For all implementations we use 8-bit VGG16 ANN and SNN (with 4 timesteps) trained on the CIFAR10 dataset. The SpikeSim\cite{moitra2023spikesim} and Neurosim\cite{chen2018neurosim}-based hardware parameters are shown in Tables \ref{tab:xbar_params} \& \ref{tab:xbar_params2} in the Appendix, respectively.}
\label{fig:snn_overview}

\end{figure*}

\textbf{von-Neumann Accelerators:}  Traditional von-Neumann AI accelerators (shown in Fig. \ref{fig:vn_imc}a), such as GPUs and TPUs, contain an array of processing elements (PE) \cite{jouppi2017datacenter, xu2018deep, chen2016eyeriss}. Each PE contains multipliers and accumulators that facilitate multi-bit MAC operations. For MAC operations, first, the weights/activation values are fetched from the off-chip DRAM memory to the input cache. Next, these values are transferred to the scratch pads of the PEs and the output is stored in the output cache and then sent back to the off-chip DRAM. During the inference of most modern deep learning networks such as ANNs and SNNs, there is a significant data exchange (in form of weights, activations and MAC outputs) between the DRAM and on-chip memories. The continual data movements and the constrained memory bandwidth contribute to the ``memory wall bottleneck" in von-Neumann architectures \cite{verma2019memory, sebastian2020memory}. Additionally, the MAC computation occurs in a cycle-to-cycle fashion. These factors degrade the throughput and energy-efficiency of von-Neumann accelerators in edge-computing scenarios. 


\textbf{IMC Dot-product Accelerators:} To overcome the "memory wall bottleneck" in von-Neumann computing, IMC (shown in Fig. \ref{fig:vn_imc}b) architectures co-locate the computation and memory units \cite{verma2019memory, sebastian2020memory}. IMC architectures feature 2D memristive crossbars, with Non-volatile Memory (NVM) devices situated at the cross-points. The NVM devices are interfaced in series with access transistors in a 1T-1R configuration to prevent sneak path currents in the crossbars \cite{yoon2018low, feng2021self}. Some NVM devices predominantly used are Phase Change Memory (PCM) \cite{burr2016recent}, Resistive Random-Access Memory (RRAM) \cite{yu2011electronic}, Spin-Torque-Transfer Magnetic RAM (STT-MRAM) \cite{wang2009spintronic} and Ferroelectric Field-effect Transistor (FeFET) \cite{jerry2017ferroelectric}. All the weights of a neural network are stored on the crossbar encoded as synaptic conductances in the NVM devices. This eliminates the weight specific data movement between memories as observed in von-Neumann architectures.

For MAC computations, the digital inputs are converted to analog voltages by the Digital-to-Analog Converter (DAC) and sent along the crossbar rows (or select-lines). These voltages get multiplied with the device conductances using Ohm's Law yielding currents which get accumulated over the crossbar column (or bit-line) according to Kirchoff's Current Law. The column currents represent the MAC operation result between inputs and weights. The Analog-to-Digital Converter (ADC) converts column currents into digital outputs. Due to the analog nature of computing, IMC architectures can facilitate highly parallel MAC computations per cycle in an energy and area-efficient manner.

\subsection{Standard Hardware Evaluation Metrics} In evaluating the efficacy of AI hardware accelerators, several key metrics are paramount. These metrics provide a quantitative basis for comparing accelerator designs and are crucial for identifying areas of improvement.

\textbf{(1) Power Consumption and Latency:} Power consumption in an accelerator comprises both dynamic and static components. Dynamic power refers to the power consumed by the accelerator during active computation, whereas static power is the power consumed at idle state. The inclusion of more hardware resources directly escalates both dynamic and static power consumption.

Latency measures the time required for an AI workload to complete its execution on the accelerator for a given input. It is a critical factor in determining the speed at which the accelerator can process data, affecting its real-time performance and user experience.

\textbf{(2) Throughput (TOPS):} As shown in Equation \ref{eq:TOPS}, throughput reflects the rate at which operations are executed per second in the accelerator. This metric is particularly relevant for AI accelerators, where an "operation" implies a multiply-and-accumulate computation.

\begin{equation}
        TOPS = \frac{Total~number~of~operations ~(in ~Tera)}{Latency}
        \label{eq:TOPS}
    \end{equation}
    
\textbf{(3) Energy-efficiency (TOPS/W):} Energy-efficiency, expressed as TOPS-Per-Watt (TOPS/W), gauges the number of operations an accelerator can perform per watt of power consumed. This metric is instrumental in assessing the sustainability and cost-effectiveness of an accelerator. It is computed as follows:

\begin{equation}
        TOPS/W = \frac{Total~number~of~operations ~(in ~Tera)}{Latency \times Power}
        \label{eq:TOPS_W}
    \end{equation}

It underscores the importance of optimizing both the computational throughput and power efficiency to enhance the overall performance of hardware accelerators.

\textbf{(4) Area-efficiency (TOPS/mm$^2$):} Area-efficiency, measured in TOPS-per-square-millimeter (TOPS/mm$^2$), evaluates the computational density of an accelerator, showcasing how effectively it utilizes its physical space to perform operations. It is computed as follows:

\begin{equation}
        TOPS/mm^2 = \frac{Total~number~of~operations ~(in ~Tera)}{Latency \times Accelerator~Area}
        \label{eq:TOPS_mm2}
    \end{equation}

It underscores the accelerator's ability to maximize its computational output relative to its size, indicating the efficiency of hardware design in terms of area utilization.

\subsection{Synergies between IMC Accelerators and SNNs}

\label{sec:synergies}
\textbf{Energy-efficiency of IMC accelerators: }Fig. \ref{fig:snn_overview}a exhaustively compares the different AI acceleration platforms used today. While GPUs and CPUs offer extensive backend support (such as CUDA\cite{luebke2008cuda} for Nvidia GPUs) for AI acceleration, their hardware architecture is fixed and not suitable for extremely low power applications ($<$ 1W). To this end, systolic accelerators such as Eyeriss \cite{chen2016eyeriss, chen2019eyeriss} (denoted as Eyeriss ANN in Fig. \ref{fig:snn_overview}a) have used ANN-centric dataflow and architecture modifications in order to achieve low power and energy-efficient acceleration. With IMC architecture (denoted as Neurosim ANN), the energy-efficiency is further improved. This is mainly attributed to the reduced weight data movement across memories and the analog dot-product operations. 



\textbf{SNNs on IMC can further energy-efficiency: }Evidently, due to the sparse and binary spike computations, SNNs can further improve the energy-efficiency and reduce power consumption in both systolic (SATA SNN \cite{yin2022sata}) and IMC accelerators (SpikeSim SNN \cite{moitra2023spikesim}) compared to ANNs. {To properly benchmark the improvements of SNNs, over ANNs, in this section, we have used SATA \cite{yin2022sata} and SpikeSim \cite{moitra2023spikesim} for SNN implementation as they closely represent the ANN-implementation architectures of Eyeriss \cite{chen2016eyeriss, chen2019eyeriss} and Neursosim \cite{chen2018neurosim}, respectively.} However, in case of SNNs implemented on systolic accelerators like SATA \cite{yin2022sata}, the SNN is still memory bottlenecked. This leads to a 13$\times$ TOPS/W improvement at 6.5$\times$ lower power compared to Eyeriss\cite{chen2016eyeriss, chen2019eyeriss} ANN. In contrast, the SNNs implemented on IMC architectures are not memory bound. Instead, IMC architectures typically suffer from the peripheral overhead of ADCs and communication circuits. Interestingly, as seen in Fig. \ref{fig:snn_overview}b, SNNs possess high activation sparsity compared to ANNs. To this end, the highly sparse binary computation in SNNs can be heavily leveraged to attain extremely low ADC precision which in turn reduces the communication overhead. Thus, SpikeSim SNN \cite{moitra2023spikesim} (with algorithmic optimizations like MINT\cite{yin2023mint} and DT-SNN \cite{li2023input} explained in Section \ref{sec:bottlenecks_and_mitigation}) yields 26$\times$ higher energy-efficiency at 4$\times$ lower power compared to Neurosim ANN. The  peripheral overhead reduction in SNN yields several synergistic benefits including 4$\times$ lower area (Fig. \ref{fig:snn_overview}c), 1.6$\times$ higher TOPS (Fig. \ref{fig:snn_overview}d) and 6.67$\times$ higher TOPS/mm$^2$ (Fig. \ref{fig:snn_overview}e) compared to Neurosim ANN \cite{chen2018neurosim}. 

\section{System-level Analyses of IMC-SNN}

\begin{figure*}
    \centering
    \includegraphics[width=1\linewidth]{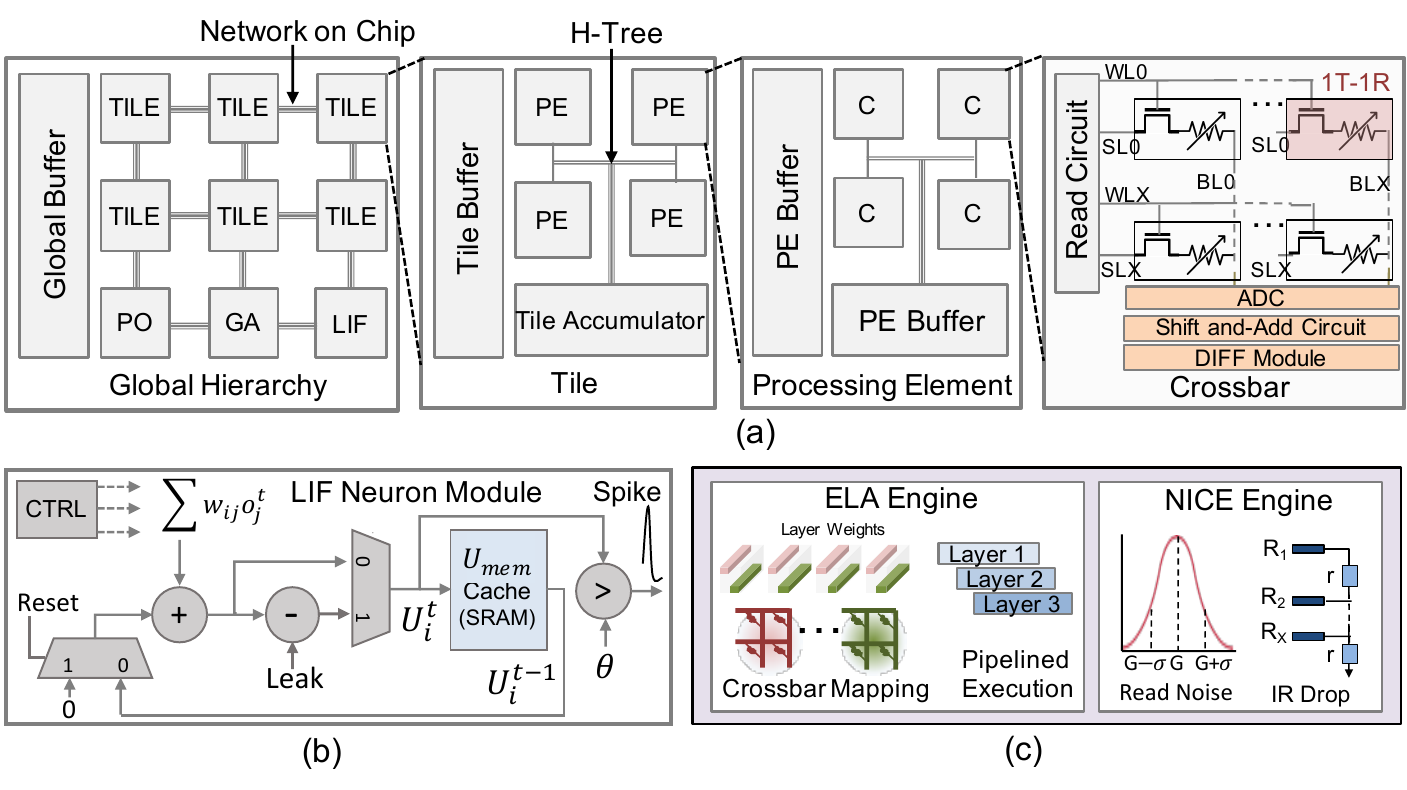}
    \caption{(a) Figure showing the hierarchical architecture of SpikeSim \cite{moitra2023spikesim} containing tiles, processing elements (PEs), and crossbars. The tiles and crossbars/PEs are connected by the network on chip (NoC) and H-Tree interconnect, respectively. The DIFF module replaces the dual crossbar approach to perform signed-MAC operations using a single crossbar. (b) The LIF neuron module stores the temporal membrane potentials in the $U_{mem}$ cache and facilitates the leaky-integrate-and-fire activation. (c) The energy-latency-area (ELA) engine and non-ideality computation engine (NICE) perform hardware-realistic energy, latency, area and accuracy evaluation during SNN inference, respectively. For SNN-IMC hardware evaluation, the SpikeSim parameters are shown in Table \ref{tab:xbar_params} in the Appendix. }
    \label{fig:spikeflow_arch}
\end{figure*}
\subsection{IMC Hardware Evaluation Platform}
\label{sec:SpikeSim}
Often, large-scale deep SNNs mandate the need for multiple crossbars integrated with numerous digital peripheral modules. For accurate system-level analyses, IMC-realistic evaluation platforms are necessary. To this end, this section highlights the extensive research performed towards large-scale ANN implementations on IMC architectures. Thereafter, it highlights the key modifications that some of the recent SNN-IMC evaluation platforms \cite{moitra2023spikesim} entail to incorporate SNNs. 

\textbf{ANN-IMC Evaluation Platforms:} In an early instance of ANN-based IMC deployment, ISAAC \cite{shafiee2016isaac} introduced a pipelined accelerator featuring on-chip embedded DRAM (eDRAM) for inter-stage data storage. The researchers conducted extensive design space exploration to determine an optimal configuration of memristor, ADCs, and eDRAM resources. In another study, MNSIM \cite{xia2017mnsim} proposed a unified framework integrating analog and digital IMC platforms for ANN implementations. Additionally, Neurosim-v1 \cite{chen2018neurosim} presented an end-to-end evaluation framework spanning device, circuit, and system levels for ANNs. The study validated simulation-based findings through post-tapeout testing, demonstrating minimal discrepancies between real and simulated outcomes \cite{lu2021neurosim}. More recent endeavors, like SIAM \cite{krishnan2021siam}, introduced IMC-based chiplet architectures tailored for ANN execution. Despite the proliferation of IMC platforms for ANNs, they lack the key SNN-specific modules such as the LIF neuron module and temporal dataflow, which are crucial for accommodating SNNs.

{\textbf{SNN-IMC Evaluation Platforms:} Over the previous years, there have been several IMC-based SNN platforms. Liu \etal \cite{liu2015spiking} proposed an analog crossbar approach for implementing feedforward and Hopfield networks, devoid of convolutional network-based dataflow. Narayanan \etal \cite{narayanan2017inxs} and Zhao \etal \cite{zhao2017analog} have constructed small feedforward SNNs trained using STDP algorithms. Bohnsting \etal \cite{bohnstingl2020accelerating} and ReSPARC \cite{ankit2017resparc}, along with Kulkarni \etal \cite{kulkarni2020chip}, have undertaken large-scale SNN deployments employing analog synapses and neurons, though these implementations lack open-source availability. To this end, SpikeSim \cite{moitra2023spikesim} proposes the first open-source end-to-end IMC hardware evaluation platform for benchmarking large-scale SNNs. This review will extensively utilize SpikeSim to perform end-to-end system-level analyses of IMC-implemented SNNs. It should be noted that, although the results are based on SpikeSim, the analyses are applicable to implementing an SNN on any IMC architecture.}

\textbf{SpikeSim- An SNN-IMC Evaluation Platform:} Similar to prior ANN works, SpikeSim \cite{moitra2023spikesim} contains a tiled hierarchical architecture containing tiles, Processing Elements (PEs) and crossbars as shown in Fig. \ref{fig:spikeflow_arch}a. The tiles are connected by a network-on-chip (NoC) interconnect, while the PEs and crossbars are connected by  H-Tree interconnects. The crossbars, PE and Tiles work in tandem to compute the MAC output at a particular timestep. Accumulators at each hierarchy add the partial sums to deliver the final MAC output. 

SpikeSim entails several architectural modifications for implementing SNNs. Firstly, the authors implement a digital neuron module that facilitates the LIF activation function. The LIF module (see Fig. \ref{fig:spikeflow_arch}b) is implemented at the global hierarchy and contains a $U_{mem}$ cache memory to store the membrane potential values over multiple timesteps. At each timestep, the membrane potential value is read from the $U_{mem}$ cache, added to the MAC output of the current timestep and written back. Secondly, the authors leverage the binary spike nature of SNNs to replace the dual-crossbar approach with a cost-efficient digital DIFF module to carry out signed MAC operations. Finally, the authors employ an SNN-specific layer-scheduled dataflow that improves the throughput and hardware utilization of IMC-implemented SNNs compared to the tick-batched dataflow \cite{narayanan2020spinalflow} used in SNN-specific systolic array accelerators. 

During inference, the SNN model weights are partitioned onto multiple crossbars and the layer-scheduled dataflow is applied. Simultaneously, SpikeSim further optimizes the floor-planning, NoC and the neuron module overhead. Following this, SpikeSim employs two engines (shown in Fig. \ref{fig:spikeflow_arch}c)- the Non-Ideality Computation Engine (NICE) and the Energy-Latency-Area (ELA) engine, to compute hardware-realistic accuracy and energy-latency-area metrics, respectively. 

\subsection{Need for System-level Analyses of IMC-SNN}

\textbf{Device innovations have been system agnostic:} Over the years, comprehensive research efforts have introduced numerous synaptic NVM devices showcasing plasticity akin to neurons in the brain \cite{van2017non, suri2011phase, la2015filamentary, boybat2018neuromorphic}. These devices aim to enable low-power unsupervised learning methods such as STDP on memristive crossbars. However, given the current scale of learning tasks, BPTT-based SNN training algorithms have become increasingly pervasive as shown in Section \ref{sec:Algorithm}. Interestingly, BPTT-trained SNNs do not require plasticity-aware synaptic devices. In fact, today's device research is geared towards achieving multi-level synaptic devices with a greater number of stable conductance states \cite{jerry2017ferroelectric}, higher On/Off ratios \cite{zhang2020low} and lower read voltages to avoid write disturbances and lower read energy during inference \cite{jao2021design, zhang2020low, raffel2022synergistic}. Concurrent studies are focused on investigating neuro-mimetic properties in emerging NVM devices for emulating analog spiking neurons \cite{stoliar2017leaky, zhang2017artificial, wang2016engineering, fang2019neuro, tuma2016stochastic, schroedter2023pseudo}. {New neuron models have also been proposed for improving SNN convergence and hardware implementations \cite{han2020rmp, shaaban2024rt}}. This research aims to enable seamless integration with analog crossbar-based dot-product engines, contributing to the development of low-power neuromorphic systems. FeFET devices having tunable hysteretic behavior and low-power switching capabilities, have shown promise in emulating the firing patterns observed in biological neurons \cite{fang2019neuro, wang2019ferroelectric}. Device researchers are also exploring the use of NVM devices like FeFETs as NVM memcapacitors for neuromorphic computing \cite{kim2023tunable, hwang2022capacitor, engeler1991capacitive}. Memcapacitive crossbars, unlike the memristive ones, perform analog dot-products in the charge-domain with low dynamic power at negligible static power dissipation. Also, the immunity of memcapacitive crossbars to sneak path currents eliminates the need for access transistors, thereby reducing design complexity and crossbar area \cite{chen2019selector}. Thus far, research at the device level has proceeded in isolation, devoid of consideration for broader system-level impact, resulting in a discernible gap in the efficient deployment of SNNs on IMCs.

\begin{table}[h!]

\Huge
    \centering
    \resizebox{\linewidth}{!}{
    \begin{tabular}{|l|ccccc|} \hline
        \multirow{2}{*}{\begin{tabular}{c} 
        \backslashbox{Device}{System}
        \end{tabular}} &  \multirow{2}{*}{\begin{tabular}{c} 
             ADC \\
             Precision
        \end{tabular}} & \multirow{2}{*}{\begin{tabular}{c}
             Acc \\
             Precision
        \end{tabular}} & \multirow{2}{*}{\begin{tabular}{c}
             Comm. \\
             BW
        \end{tabular}} & \multirow{2}{*}{\begin{tabular}{c}
             Xbar \\
             Count
        \end{tabular}} & \multirow{2}{*}{\begin{tabular}{c}
             Buffer \\
             BW
        \end{tabular}} \\
        & & & & & \\ \hline
        
        Dev. Precision $\uparrow$ & $\uparrow$ & $\uparrow$ & $\uparrow$ & \textcolor{red}{$\downarrow$} & $\uparrow$ \\ \hline
            
         On Resistance $\uparrow$ & $\times$ & $\times$ & \textcolor{red}{$\downarrow$} & $\times$ & \textcolor{red}{$\downarrow$} \\ \hline

         On/Off Ratio  $\uparrow$ & $\times$ & $\times$ & \textcolor{red}{$\downarrow$} & $\times$ & \textcolor{red}{$\downarrow$} \\ \hline

         Read Voltage ~$\uparrow$ & $\times$ & $\times$ & $\uparrow$ & $\times$ & $\uparrow$ \\ \hline
    \end{tabular}}
    \caption{Table showing the co-dependency between different device and system parameters. Non co-dependent device and system parameters are denoted by $\times$. Dev., Xbar, Acc., Comm. BW refer to device, crossbar, accumulator, communication bandwidth, respectively.}
    \label{tab:codependency_table}
    \vspace{-4mm}
\end{table}
\textbf{Co-dependence among Device and System Parameters:} Table \ref{tab:codependency_table} shows the system-level parameters that are co-dependent on the different NVM device parameters. Increasing the device precision (number of stable conductance states) reduces the crossbar count required for IMC mapping. However, it also increases the ADC precision to process larger crossbar currents and, in turn, the bandwidth requirements of the buffers and the communication circuits are high (Refer Fig. \ref{fig:device_precision} for details). Similarly, increasing the device On resistances and the On/Off ratios increases read parallelism (the number of crossbar rows read in parallel). However, it slows down the system's frequency of operation (Refer Fig. \ref{fig:ni_results}d for details). Lower operation frequency will also lower the demand for communication and buffer bandwidths. Finally, increasing parameters such as the read voltage (considering no write disturbances) of the NVM devices increases the system's frequency of operation, and hence the requirement for higher interconnect and buffer bandwidths. Therefore, it is imperative to grasp the co-dependencies among device, circuit, and system-level parameters to fully analyze the energy-latency-area-accuracy landscape of SNNs implemented on IMC architectures.

\subsection{SNN System-level Bottlenecks and Mitigation Strategies}
\label{sec:bottlenecks_and_mitigation}
The intrinsic energy-efficiency of SNNs may be compromised without a thorough understanding of hardware bottlenecks. This section delineates the primary obstacles hindering the efficient integration of SNNs on IMC architectures.

\subsubsection{\textbf{The LIF Neuron Module}}
\label{sec:lif_overhead}
\begin{figure}[h!]
    \centering
    \includegraphics[width=0.8\linewidth]{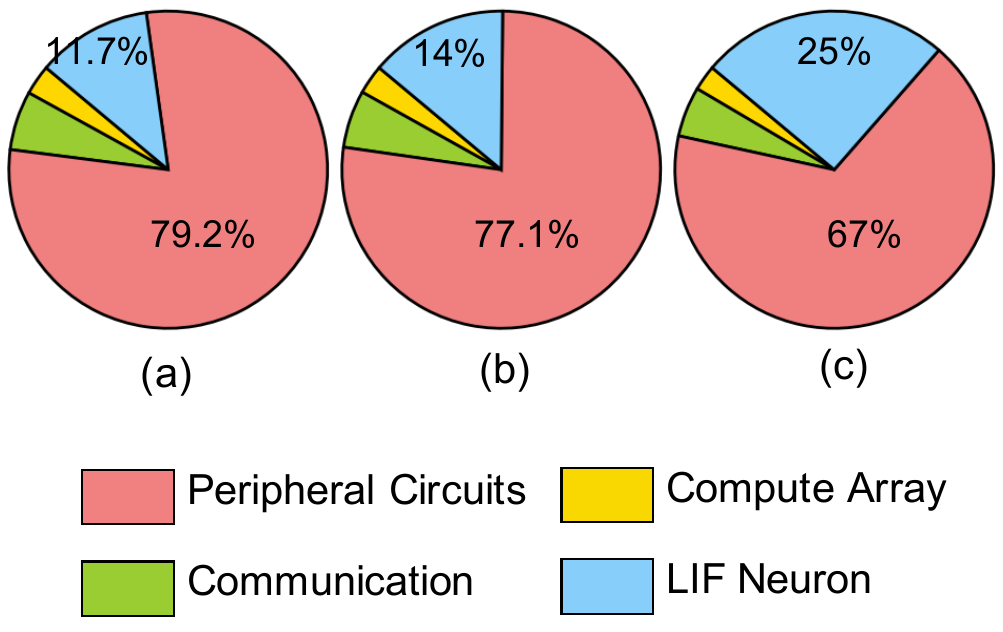}
    \caption{Area distribution of 8-bit VGG16 SNN on SpikeSim (with hardware parameters shown in Table \ref{tab:xbar_params} in the Appendix) with (a) CIFAR10 (b) Caltech-101 and (c) TinyImagenet datasets. Peripheral circuits include ADCs, accumulators, buffers, read circuit, shift-and-add circuit and DIFF module. Communication involves NoC and H-Tree circuits. Compute array consists of all the 2D 1T-1R crossbar arrays excluding the peripheral circuits. {These trends will remain consistent irrespective of the SNN-IMC platform used, as they are determined by the memory cell area and the dataset feature size.}}
    \label{fig:lif_distribution}
\end{figure}

Fig. \ref{fig:lif_distribution}a-c, show the area distribution across different modules of the SpikeSim platform for an 8-bit VGG16 SNN trained on CIFAR10\cite{cifar10} (image dimensions of 32$\times$32) , Caltech-101\cite{li_andreeto_ranzato_perona_2022} (image dimensions of 48$\times$48) and TinyImagenet\cite{le2015tiny} (image dimensions of 64$\times$64)  datasets, respectively. Evidently, due to the large $U_{mem}$ (refer Fig. \ref{fig:spikeflow_arch}b) cache memory, the LIF neuronal module contributes 11.7\%-25\% towards the overall chip area. The LIF module thus, poses a bottleneck when implementing SNNs trained on large datasets like ImageNet \cite{deng2009imagenet} with image dimensions (224$\times$224 and 384$\times$384) on IMC architectures. Interestingly, the LIF module requires $>1000\times$ higher on-chip area compared to ReLU module in ANNs that merely requires a comparator.

\textbf{Co-design based Mitigation Strategies:} In SpikeSim \cite{moitra2023spikesim}, the authors propose a simple approach of channel scaling wherein, the number of output channels in the first layer of the network are scaled down yielding 2$\times$ reduction in LIF module area. In MINT \cite{yin2023mint}, the authors apply sophisticated weight and membrane potential quantization-aware SNN training to reduce the LIF overhead. MINT is able to reduce the weight and membrane potential precision as low as 2 bits while maintaining iso-accuracy with SNN trained on FP32 precision. Additionally, another recent work \cite{kim2023sharing} performed sharing of LIF membrane potentials over multiple SNN layers to reduce the LIF memory overhead. The authors perform inter-layer and intra-layer membrane potential sharing to achieve over 4$\times$ reduction in the LIF memory area at iso-accuracy.  


\textbf{Device Research for LIF Area Mitigation:} Device researchers are exploring novel neuromimetic devices emulating biological neuronal functionalities. Recent works have leveraged the fast and low-power switching dynamics of a FeFET-based relaxation oscillator configuration to generate biological spiking patterns at area-efficient form factors \cite{fang2019neuro, wang2019ferroelectric}. In another work Mohanan \etal \cite{mohanan2024optimization} used nanoporous graphene-based memristive devices to compactly emulate LIF neurons in current SNN workloads showing threshold control, leaky integration and reset behaviors. The spiking activity of the LIF neuron is tunable by varying various circuit and device parameters, allowing it to cover a broad frequency spectrum \cite{mohanan2024optimization}. Likewise, Zhou \etal \cite{zhou2022gradient} have proposed a compact RRAM-based LIF neuron circuit closely integrated with analog RRAM crossbars. This provided a unified path to carry out dot-products and LIF activation functionalities in the analog domain.
However, given the large number of spatial channels required by large-scale BPTT-trained SNN models, directly integrating the analog LIF neurons with the crossbars remains an unsolved problem.

\subsubsection{\textbf{Temporal Computation in SNNs}}
\label{sec:ts_overhead}
\begin{figure}[h!]
    \centering
    \includegraphics[width=0.8\linewidth]{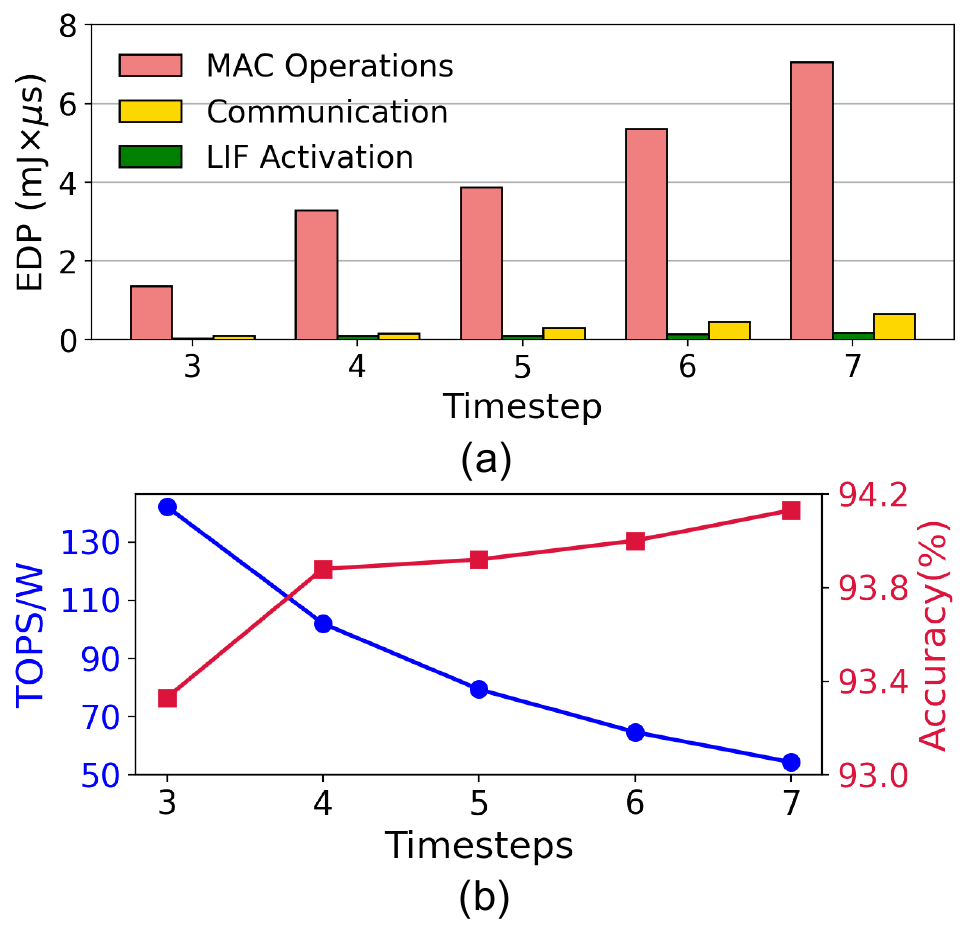}
    \caption{Figure showing the (a) trends in energy-delay product (EDP) across the MAC operations, LIF activations and communication circuits. (b) Trends of TOPS/W and accuracy with increasing timesteps. Results correspond to 8-bit VGG16 SNN implemented with SpikeSim trained on the CIFAR10 dataset. SpikeSim parameters for SNN implementation are shown in Table \ref{tab:xbar_params} in the Appendix.}
    \label{fig:ts_overhead}
\end{figure}
As SNNs process data over multiple timesteps, an increase in timesteps linearly escalates the energy-delay product (EDP) across MAC, communication and LIF activation operations (Fig. \ref{fig:ts_overhead}a). Interestingly, the crossbar compute arrays, digital peripherals and the communication circuits get activated multiple times in a particular timestep in order to compute the weighted summation output. In contrast, the LIF activation is performed once every timestep. Therefore, the MAC and communication operations significantly contribute to the EDP (80\% of the overall EDP). Consequently, reducing the number of timesteps can significantly enhance efficiency. However, Fig. \ref{fig:ts_overhead}b illustrates a trade-off between timesteps, energy-efficiency and accuracy. While reducing timesteps improves energy-efficiency, it also leads to lower SNN accuracy. Therefore, development of effective algorithms exploiting spatio-temporal complexity is imperative. {Note that while Fig. \ref{fig:ts_overhead} uses SpikeSim for evaluation, the timestep is an intrinsic parameter of the SNN algorithm. Consequently, the linear increase in EDP with timesteps will remain consistent regardless of the IMC platform.}


\textbf{Co-design based Timestep Minimization Strategies: } Over the years, training algorithms such as BNTT \cite{kim2021revisiting}, along with neuromorphic data augmentation techniques \cite{li2022neuromorphic} and encoding methods such as direct encoding \cite{kim2022rate}, have effectively exploited the spatio-temporal complexity of SNNs resulting in a drastic reduction (of the order 10$\times$) in the number of timesteps. A more recent approach by Li \etal \cite{li2023input} called DT-SNN leverages the difficulty of the input images to scale the number of timesteps in the SNN. During training, the authors use a joint training loss to train an SNN on different count of timesteps. During inference, the authors use an entropy metric to determine the confidence of prediction per timestep. An image with lower entropy is deemed easy and inferred at an early timestep and difficult images with higher entropy are inferred at latter timesteps. The authors of DT-SNN achieve an overall 81\% EDP reduction, with iso or higher accuracy than a standard SNN using fixed number of timesteps for inference across all inputs .

\subsubsection{Vulnerability towards IMC Non-idealities}

\begin{figure}[t]
    \centering
    \includegraphics[width=0.9\linewidth]{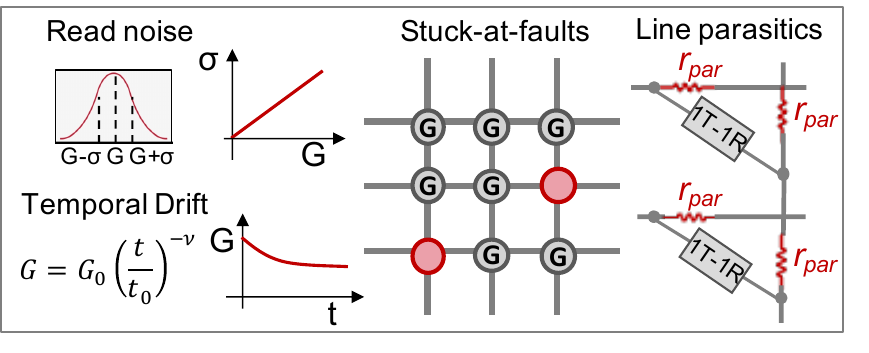}
    \caption{Non-idealities impacting inference on NVM crossbars. The memristive device-level non-idealities are read noise \cite{agarwal2016resistive, veksler2013random, nandakumar2018phase, sun2019impact}, temporal conductance drift \cite{nandakumar2018phase, chen2018reliability} and stuck-at-faults \cite{yeo2019stuck, zhang2019handling}. The parasitic resistances \cite{zhang2020mitigate, jain2020rxnn} of the metal lines in the crossbars ($r_{par}$) and the transistor non-linearities in the 1T-1R synapses \cite{bhattacharjee2021neat} comprise the circuit-level non-idealities.}
    \label{fig:ni_imc}
\end{figure}
\begin{figure*}[t]
    \centering
    \includegraphics[width=1\linewidth]{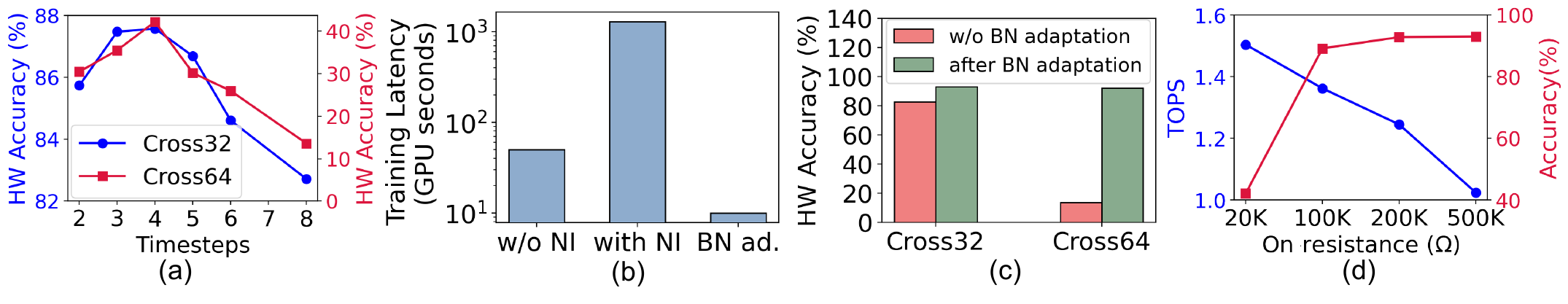}
    \caption{(a) Plot showing the impact of crossbar non-idealities on SNN inference accuracy.  We consider the effect of stray IR drops due to resistive non-idealities ($r_{par} = 5\Omega$) as well as stochastic Gaussian read noise ($\sigma=0.1$) in the RRAM devices. (b) Plot showing the GPU training latency per iteration for non-ideality-aware training (with NI) compared against standard GPU training (without non-ideality or w/o NI) and training-less noise-aware BN adaptation (BN ad.). Evaluations are performed on the Nvidia RTX2080Ti GPU. (c) Plot showing the effectiveness of noise-aware BN adaptation in mitigating crossbar non-idealities. (d) Plot showing the trend of system-level TOPS and non-ideal accuracy by varying On resistance of the synaptic RRAM devices. For a,c and d, evaluations are performed on an 8-bit VGG16 SNN implemented on SpikeSim using parameters shown in Table \ref{tab:xbar_params} in the Appendix. {The accuracy is affected by the device-level parameters and therefore will show similar trends irrespective of the SNN-IMC evaluation platform used.} Cross32 and Cross64 denote crossbars of sizes 32$\times$32 and 64$\times$64, respectively. All evaluations are performed on the CIFAR10 dataset.}

    \label{fig:ni_results}
\end{figure*}
The practical implementation of NVM devices is constrained by finite conductance levels, limited On/Off resistances, and inherent non-idealities that can adversely affect the inference accuracy of AI workloads \cite{chen2017neurosim+, chen2018neurosim, wan202033, moitra2023spikesim}. Depending upon the origin, these non-idealities can be classified into device and circuit non-idealities as shown in Fig. \ref{fig:ni_imc}.

\textbf{(1) Device Non-idealities:} Stochastic read noise predominantly originates from random telegraphic noise, flicker noise ($1/f$ noise), and thermal noise in the NVM synapses \cite{agarwal2016resistive, veksler2013random, nandakumar2018phase, sun2019impact}. Read noise is modelled as a Gaussian distribution around the programmed conductance during each read cycle, with a standard deviation ($\sigma$) increasing with the NVM conductance \cite{agarwal2016resistive, nandakumar2018phase, sun2019impact}. Structural relaxation within NVM devices over time leads to another non-ideality called {temporal drift}, influencing the retention of programmed conductance in crossbars \cite{nandakumar2018phase, chen2018reliability}. A popular model describing the temporal conductance drift is given as $G(t) = G_0*(t/t_0)^{-\nu}$, where $G_0$ represents the initially programmed conductance at time $t_0$, and $\nu$ denotes the drift coefficient. {Stuck-at-fault} is another non-ideality arising from fabrication defects or extensive crossbar utilization.  Stuck-at-faults manifest into fixated NVM synapses in crossbars (to set or reset states), rendering them non-programmable \cite{yeo2019stuck, zhang2019handling}. Despite the inherent robustness of NVM crossbars to variations, stuck-at-faults can significantly degrade the performance of ANN/SNN workloads.

\textbf{(b) Circuit-level Non-idealities:} It includes the {parasitic resistances} in the crossbar metal lines denoted as $r_{par}$. During analog dot-product operations, the interconnect parasitics lead to stray IR drops, causing the output currents to deviate from their ideal values and resulting in substantial accuracy losses \cite{zhang2020mitigate, jain2020rxnn, bhattacharjee2022examining}. Balancing crossbar size to improve parallelism during inference, while restraining the impact of resistive non-idealities, becomes a delicate trade-off. Furthermore, the presence of access transistors in series with NVM devices in 1T-1R synapses is crucial for eliminating sneak paths and incorrect programming of the NVM devices \cite{yoon2018low, feng2021self}. Nonetheless, it introduces {1T-1R non-linearities} arising from the non-linear I-V characteristics of the access transistors \cite{bhattacharjee2021neat}.

\textbf{Effect of IMC Non-ideality on SNN:} For an 8-bit VGG16 SNN model trained with the CIFAR10 dataset, the impact of parasitic resistances and stochastic read noise on the hardware inference accuracy is shown in Fig. \ref{fig:ni_results}a. At higher timesteps (timesteps $\geq$ 4), we find the non-ideal inference accuracy declines dramatically, owing to significant non-ideality error accumulation over multiple timesteps of computations \cite{bhattacharjee2022examining}. 

\textbf{Non-ideality-aware Training of SNNs:} Ensuring  robustness to crossbar non-idealities involves iterative offline-training of SNN models with noise injection using hardware-realistic noise models \cite{liu2015vortex, roy2021txsim, chakraborty2020geniex, charan2020accurate, dampfhoffer2023improving}. Recently, the AIHWKit toolkit from IBM offers statistical empirical models for emulating device-level and parasistic resistive non-idealities for PCM crossbars in PyTorch \cite{rasch2021flexible, rasch2023hardware}. AIHWKit-based noise-aware training has demonstrated state-of-the-art hardware accuracies across diverse tasks, spanning computer vision to natural language processing. However, non-ideality-aware training is not scalable to today's large-scale SNN models. This is due to the bottleneck of noise-injection, especially for the input voltage-dependent resistive parasitic non-idealities, which escalates training latency. Additionally, SNNs incur high training costs due to computations across multiple time-steps. This has been depicted in Fig. \ref{fig:ni_results}b where, an iteration of non-ideality-aware training increases the training latency by greater than an order of magnitude compared to standard training. This underscores the need for training-less methods for non-ideality mitigation. 

\textbf{Training-less Non-ideality Mitigation Strategies:} Recent methods have proposed transformations on NVM conductances during mapping onto crossbars, increasing the proportion of low conductance synapses to mitigate crossbar non-idealities \cite{bhattacharjee2021neat, bhattacharjee2023switchx, bhattacharjee2024clipformer}. Based on this principle, the NICE engine in the SpikeSim framework shows significantly improved SNN inference accuracies on non-ideal crossbars with no additional hardware costs \cite{moitra2023spikesim}. In addition, a recent work has shown that simple noise-aware adaptation of the batch-normalization (BN) parameters of a BPTT-trained SNN can fully recover the inference accuracy lost due to the non-idealities \cite{bhattacharjee2022examining}. This is corroborated in Fig.\ref{fig:ni_results}c across crossbar sizes of 32$\times$32 and 64$\times$64. Noise-aware BN adaptation is a fully weight-static approach, implying that NVM synapses need not be re-programmed or reconfigured during inference to mitigate non-idealities \cite{bhattacharjee2023examining}. Noise-aware BN adaptation incurs nearly an order of magnitude lower latency than one epoch of standard SNN training (see Fig. \ref{fig:ni_results}b). 

\textbf{Choice of NVM Device for Non-ideality Mitigation:} While RRAMs and PCMs are extensively studied for multi-level crossbar synapses, their susceptibility to read noise is a concern. In contrast, FeFET-based synapses, with increased CMOS-compatibility and high On/Off ratios (>100), show promise in reducing read noise \cite{zhang2020low}. PCMs have exhibited high retention capabilities ($>10$ years) for temporal drift \cite{pirovano2004reliability, nandakumar2018phase, le2020phase}, while FeFETs show poorer retention ($\sim10^3-10^4s$) due to polarization degradation from charge traps, defects, and oxide breakdown \cite{huang2011retention}. To minimize stray IR drops, synapses with high On resistance (typically $>100k\Omega$) are favoured \cite{roy2021txsim}. 

However, excessively high On resistances diminish the crossbar currents, impacting the readout by sense amplifiers or ADCs \cite{jao2021design}. This is corroborated in Fig. \ref{fig:ni_results}d, where increasing On resistance of the synaptic devices leads to higher accuracy for the VGG16 SNN on 64$\times$64 crossbars, while reducing the TOPS at the system-level. The reduction in TOPS manifests from the reduced crossbar currents driving the ADCs. Furthermore, with research on memcapacitive NVM devices gaining momentum \cite{kim2023tunable, hwang2022capacitor, engeler1991capacitive}, it is noteworthy that the crossbars operating in the charge-domain eliminate the impact of circuit-level non-idealities such as stray IR drops and 1T-1R non-linearities \cite{kim2023tunable}.

\section{Discussion and Future Directions}

\subsection{Does IMC Need Very High Device Precisions?}

\begin{figure}[h!]
        \centering
        \includegraphics[width=0.8\linewidth]{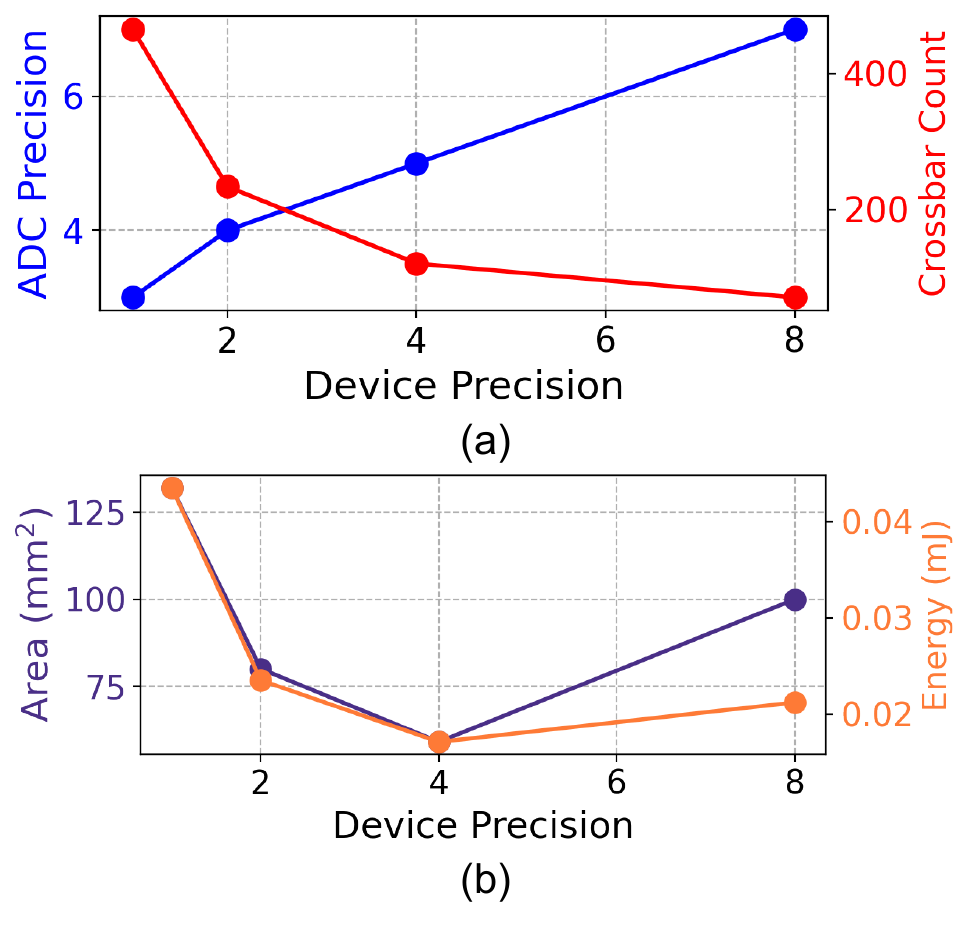}
        \caption{Figure showing the (a) trend of required ADC precision and crossbar count with increasing device precision. (b) the trend in area and energy upon increasing device precision. All evaluations are performed on SpikeSim with 8-bit VGG16 SNN implemented using hardware parameters shown in Table \ref{tab:xbar_params} in the Appendix. }
        \label{fig:device_precision}
\end{figure}

\begin{figure*}[t]
    \centering
    \includegraphics[width=.9\linewidth]{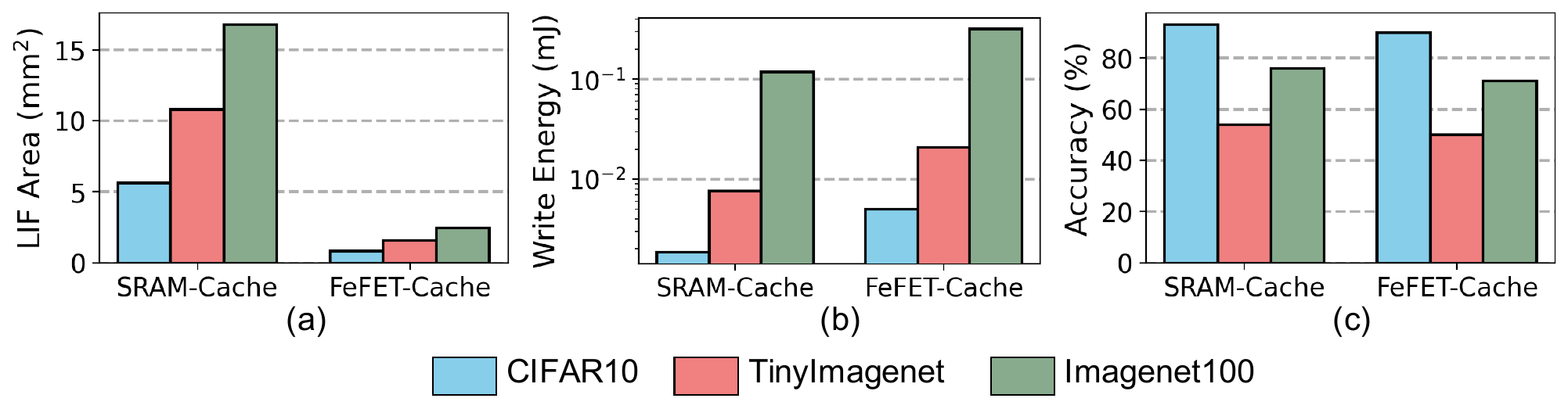}
    \caption{Figure comparing the (a) LIF module area (b) Write Energy (c) Accuracy of SpikeSim-implemented SNNs implemented with SRAM and FeFET-based $U_{mem}$ cache inside the LIF module. Results are shown across the CIFAR10, TinyImagenet and Imagenet100 datasets for an 8-bit VGG16 SpikeSim-implemented SNN with parameters shown in Table \ref{tab:xbar_params} in the Appendix. {These trends will remain consistent irrespective of the SNN-IMC platform used, as they are determined by the memory cell area and the dataset feature size.}}
    \label{fig:fefet_lif}
    \vspace{-3mm}
\end{figure*}
    
The device community has always focused on targeting higher number of stable conductance values in NVM devices without considering broader system-level implications. One might naturally assume that enhancing the precision of NVM devices will reduce the number of crossbars (and their associated peripherals) needed to implement SNN layers. However, at higher device precisions, the ADC precision needs to increase (and hence, the ADC area and energy) to avoid quantization-errors in the accumulated column currents, resulting in an expanded area and energy at the system level. The co-dependence of device precision with ADC precision is illustrated in Fig. \ref{fig:device_precision}a. For the 8-bit VGG16 SNN model, the optimal NVM device and ADC precisions are found to be 4 bits and 5 bits, respectively. This yields the best energy and area expenditures at the system-level as shown in Fig. \ref{fig:device_precision}b. Therefore, to attain considerable energy and area-efficiency, large device precision is not paramount. {It must be noted that these trends are IMC platform agnostic as they are solely governed by the device precision.}


\subsection{FeFETs as a Promising Device for $U_{mem}$ Cache} In light of the LIF area overhead discussed in Section \ref{sec:lif_overhead}, utilizing an NVM device like FeFET for constructing $U_{mem}$
cache could drastically reduce the LIF area by upto $7\times$ compared to the traditional SRAM  cache (see Fig. \ref{fig:fefet_lif}a). However, as illustrated in Fig. \ref{fig:fefet_lif}b, current FeFET technology necessitates multiple write cycles for programming (Refer Section \ref{sec:online_learning} for details on writing into NVM devices), leading to $3\times$ greater write energy than that of SRAM caches. This increased write energy stems from the need to perform $U_{mem}$ write operations over multiple timesteps during SNN inference. Despite FeFETs showing superior noise-resilience compared to RRAMs and PCMs, FeFETs continue to display read and write variabilities, potentially decreasing the SNN accuracy by $3-4\%$ (Fig. \ref{fig:fefet_lif}c) across a range of datasets. Note, the relatively short retention time of FeFETs ($\sim10^3-10^4s$) is unlikely to pose concerns given that the LIF cache is updated at a significantly higher frequency, ranging from tens to hundreds of MHz. 

\subsection{Opportunities for IMC-SNNs in Online Learning}
\label{sec:online_learning}
\begin{figure}[h!]
    \centering
    \includegraphics[width=0.9\linewidth]{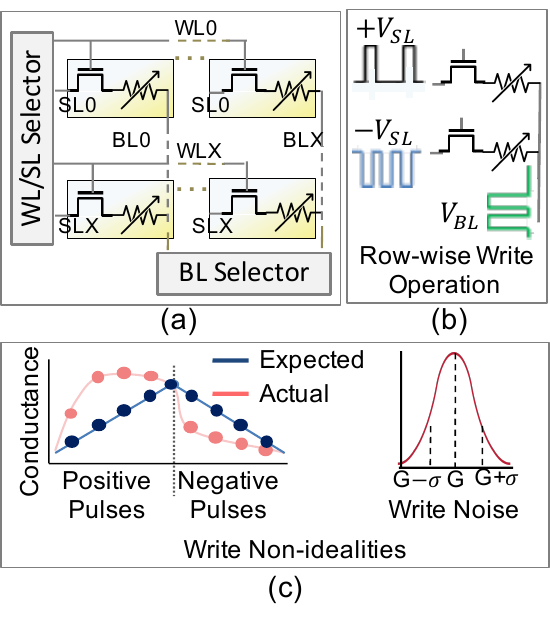}
    \caption{Figure showing (a) the essential circuits required to facilitate weight updates in IMC architectures; (b) costly NVM write operation over multiple write pulses \cite{marinella2018multiscale, chen2018neurosim}; (c) the write non-idealities \cite{chen2018neurosim, roy2021txsim} such as asymmetric weight update and write noise.}
    \label{fig:online_learn}
    \vspace{-3mm}
\end{figure}



In the recent years, there has been a growing interest in online learning on edge devices \cite{chen2019deep}. Data privacy concerns make learning on edge devices imperative because it allows sensitive or personal information to be processed directly on-device, without sending it across the internet or to centralized servers. Furthermore, for applications requiring real-time or near-real-time responses, such as autonomous vehicles or emergency response systems, online learning can save huge communication latency and bandwidth as data does not need to be communicated back and forth across the internet \cite{liu2019edge, chen2019deep}. The contemporary SNNs demonstrate ability to make accurate predictions with minimal temporal samples. They are particularly suited for edge devices due to their ability to operate at low power and their efficiency in handling highly sparse time-series data, which is common in real-world sensory inputs \cite{zhou2022gradient, lobo2020spiking}. As discussed in Section \ref{sec:synergies}, the strong synergies between SNNs and IMC crossbars, particularly the reduced communication and ADC overheads to process binary and sparse spike data, shows potential for employing SNNs on IMC crossbars for online learning.

\textbf{Device Challenges towards SNN Online Learning:} In online training, writing into NVM devices involves selecting specific synapses by applying pulses across rows (select-lines or SLs) and columns (bit-lines or BLs) (see Fig. \ref{fig:online_learn}a), followed by modulating voltage or current to adjust the synaptic conductance \cite{chen2018neurosim}.  Each write cycle can degrade the NVM device's material, affecting its lifespan, making high-endurance devices preferable. Additionally, as programming each device requires multiple pulses, write operations are delay and energy-intensive \cite{marinella2018multiscale}. This is demonstrated in Fig. \ref{fig:online_learn}a-b. Write challenges also stem from the stochastic write noise and asymmetric conductance updates in the NVM devices (see Fig. \ref{fig:online_learn}c), which, although negligible during inference, significantly impact weight re-programming during online learning \cite{marinella2018multiscale, chen2018neurosim, roy2021txsim}. These non-idealities necessitate repeated write operations to achieve the desired conductance level, affecting the energy, speed, and device endurance. 

\textbf{Hardware Requisites for Online Learning:} Write noise mitigation strategies, in general, include error correction codes \cite{niu2012low}, write verification-and-retry mechanisms \cite{li202240}, and structural advancements in the NVM devices \cite{jao2021design, raffel2022synergistic}. FeFETs are less susceptible to write noise compared to RRAMs \& PCMs, owing to deterministic polarization-switching at low voltages. However, FeFETs in general show limited endurance ($\sim10^4-10^{10}$ cycles) owing to mobility degradation and charge trapping phenomena \cite{duan2022impact, pesic2018deconvoluting, tan2021ferroelectric, ali2018high, yurchuk2014origin}. The low endurance of FeFETs can become problematic during online learning as the weights of the SNN need to be updated frequently. Furthermore, RRAMs \& PCMs rely on filament formation mechanism and high thermal energy for state change, respectively, leading to high write energy and  latency. In contrast, FeFETs offer substantial write energy and latency reductions due to the rapid and low-voltage switching of ferroelectric layers.


\textbf{Hardware-friendly Online Learning Paradigms}: Current BPTT-based algorithms entail huge memory and computational costs for facilitating backpropagation on hardware over multiple timesteps \cite{yin2022sata}. The ability to update model parameters locally and independently at each layer is important for online and continual learning paradigms. This is crucial for applications that require the model to adapt continuously to new data without the need for re-training from scratch. To this end, local gradient-based learning methods, such as Direct Feedback Alignment (DFA) \cite{nokland2016direct}, show great promise in reducing training latency and improving TOPS/W at the system-level  over traditional backpropagation\cite{lillicrap2016random, crafton2019local, lu2020accelerated}. {Additionally, emerging learning algorithms exploiting the eligibility traces in SNNs can achieve bioplausible \cite{bellec2019eligibility} and memory-efficient online learning at the edge \cite{frenkel2022reckon}.}



\subsection{Need for Layer-specific Peripheral Circuit Co-optimization}

So far, all optimizations that have taken place have been implemented homogeneously across different SNN layers, regardless of the layer-specific computational complexity. However, it is important to note that different layers have different compute complexity and therefore will require specific device-circuit-system and algorithmic parameter optimization. Recent works have proposed layer-specific device \cite{bhattacharjee2023hyde} and peripheral circuit parameters \cite{moitra2023xpert} to obtain optimal energy \& area-efficiencies. To optimize the communication overhead, work by Krishnan \etal \cite{krishnan2020interconnect} have proposed layer-specific tile sizes to minimize inter-tile communications. Here, it is important to highlight that despite the two-dimensional integration using NoCs in a typical silicon fabrication process, the on-chip connectivity still falls short of the three-dimensional connectivity observed in the brain \cite{ulloa2016role}. Consequently, more recently, 3D crossbar-based IMC architectures \cite{wang2023benchmarking, murali2020heterogeneous} have emerged as a viable solution to address this communication bottleneck. Nevertheless, all these studies have primarily focused on optimizations within the ANN domain, underscoring the importance of conducting layer-specific optimizations tailored for SNNs.

\section{Conclusion}
The review delineates the pivotal synergies between SNNs and IMC architectures, showcasing their efficacy in ultra-low-power edge computing scenarios. SNNs are being actively used for various commercial applications requiring extensive academic studies across multiple application spaces. To achieve optimal low-power edge implementations, the review motivates system-level analyses by considering the co-dependencies between algorithm, device, circuit and system parameters. Furthermore, we point out the bottlenecks at the system level that arise from implementing SNNs on IMC architectures due to NVM device limitations. To this end, our review delves into several device, circuit and system-aware co-design-based strategies that have been developed to overcome the inherent bottlenecks. Finally, we emphasize on future device research landscape to facilitate energy-efficient IMC-SNN deployment with key focus on online learning, emerging neuronal devices, and effective design-space co-exploration.

\section*{Acknowledgement}
\noindent This work was supported in part by CoCoSys, a JUMP2.0 center sponsored by DARPA and SRC, the National Science Foundation (CAREER Award, Grant \#2312366, Grant \#2318152), TII (Abu Dhabi), and the DoE MMICC center SEA-CROGS (Award \#DE-SC0023198)

\section*{Appendix}
\label{sec:appendix}
\begin{table}[h!]
    \centering
    \caption{Table with values of various circuit \& device parameters used for SpikeSim\cite{moitra2023spikesim} evaluation unless otherwise mentioned.}
    \resizebox{.7\linewidth}{!}{
    \begin{tabular}{|l|c|} \hline
       \multicolumn{2}{|c|}{\textbf{SpikeSim Evaluation Parameters}} \\ \hline

       Technology & 65nm CMOS \\ \hline
       NoC Topology & Mesh \\ \hline
       NoC Width & 32 bits \\ \hline
       Crossbar Size & 64$\times$64 \\ \hline
       
       Clock Frequency & 250 MHz \\ \hline 
       
       Crossbars/PE & 9 \\ \hline
       PE/Tile & 8 \\ \hline

       Buffer Sizes & Global-20KB, Tile-10KB, PE-5KB \\ \hline

       Read Voltage & 0.1V \\ \hline
       
       Device & RRAM \\ \hline
       Device Precision & 4 bits \\ \hline
       On/Off Ratio & 10 ($R_{on}=$20k$\Omega$)\\ \hline
    \end{tabular}}
    
    \label{tab:xbar_params}
\end{table}

\begin{table}[h!]
    \centering
    \caption{Table with values of various circuit \& device parameters used for Neurosim\cite{chen2018neurosim} evaluation unless otherwise mentioned.}
    \resizebox{.7\linewidth}{!}{
    \begin{tabular}{|l|c|} \hline
       \multicolumn{2}{|c|}{\textbf{Neurosim Evaluation Parameters}} \\ \hline

       Technology & 65nm CMOS \\ \hline
    
       Crossbar Size & 64$\times$64 \\ \hline
       
       Clock Frequency & 250 MHz \\ \hline 
       
       Crossbars/PE & 9 \\ \hline
       PE/Tile & 8 \\ \hline

       Buffer Sizes & Global-20KB, Tile-10KB, PE-5KB \\ \hline

       Read Voltage & 0.1V \\ \hline
       
       Device & RRAM \\ \hline
       Device Precision & 4 bits \\ \hline
       On/Off Ratio & 10 ($R_{on}=$20k$\Omega$)\\ \hline

    \end{tabular}}
    
    \label{tab:xbar_params2}
\end{table}

\section{References}
\bibliography{APR}

\end{document}